\documentclass[a4paper,fleqn]{cas-dc}
\usepackage[numbers,sort&compress]{natbib} 
\graphicspath{{figures/}} 
\usepackage{graphicx}
\usepackage[caption=false,font=footnotesize,labelfont=rm,textfont=rm]{subfig}
\usepackage{bm} 
\usepackage{hyperref} 
\usepackage{amsmath}
\usepackage{amsthm,amssymb} 
\usepackage{xcolor}
\usepackage{soul}
\usepackage{enumitem}
\usepackage{booktabs}
\usepackage{tabularx}
\usepackage{array}
\usepackage{makecell}
\usepackage{multirow}
\setlist[itemize]{leftmargin=1.2em,itemsep=0pt,topsep=2pt,parsep=0pt,partopsep=0pt,label=\scriptsize$\bullet$}

\usepackage[caption=false,font=footnotesize,labelfont=rm,textfont=rm]{subfig}

\makeatletter
\renewcommand{\figurename}{Fig.}
\renewcommand{\tablename}{Table}

\renewcommand{\fnum@figure}{{\fontfamily{ptm}\selectfont\small\bfseries \figurename~\thefigure.}}
\renewcommand{\fnum@table}{{\fontfamily{ptm}\selectfont\small\bfseries \tablename~\thetable}}

\ExplSyntaxOn

\cs_set:Npn \__make_tbl_caption:nn #1#2
{
  \l_tbl_align_tl
  \skip_vertical:N \l_tbl_abovecap_skip
  {
    \parbox{\dimexpr(\l_tbl_width_dim)}
    {
      \rightskip=0pt
      #1\par
      {\fontfamily{ptm}\selectfont\footnotesize #2}\par\vskip4pt
    }
  }
  \skip_vertical:N \l_tbl_belowcap_skip
}

\cs_set:Npn \__make_fig_caption:nn #1#2
{
  \l_fig_align_tl
  \skip_vertical:N \l_fig_abovecap_skip
  \setbox\cascaptionbox=\hbox
  {
    #1\hspace{0.35em}
    {\fontfamily{ptm}\selectfont\footnotesize #2}
  }
  \ifdim\the\wd\cascaptionbox<\dim_use:N \l_fig_width_dim\relax
    \parbox{\l_fig_width_dim}
    {
      \unskip\ignorespaces\hfil
      #1\hspace{0.35em}
      {\fontfamily{ptm}\selectfont\footnotesize #2}
      \hfil\par
    }
  \else
    \parbox{\l_fig_width_dim}
    {
      \rightskip=0pt\unskip\ignorespaces
      #1\hspace{0.35em}
      {\fontfamily{ptm}\selectfont\footnotesize #2}\par
    }
  \fi
  \skip_vertical:N \l_fig_belowcap_skip
}

\ExplSyntaxOff
\makeatother

\def\tsc#1{\csdef{#1}{\textsc{\lowercase{#1}}\xspace}}
\tsc{WGM}
\tsc{QE}

\begin{document}
\let\WriteBookmarks\relax
\def\floatpagepagefraction{1}
\def\textpagefraction{.001}
\let\printorcid\relax 

\shorttitle{Stage-Aware and Roughness-Constrained Diffusion Policy for Multi-Stage Robotic Polishing}    


\shortauthors{Shuai \textit{et al}}  

\title [mode = title]{Stage-Aware and Roughness-Constrained Diffusion Policy for Multi-Stage Robotic Polishing}  



%

\author[1]{Shuai Ke}[style=chinese]
\ead{keshuai@hust.edu.cn} 

\author[1]{Jiexin Zhang}[style=chinese]
\ead{zhangjiexin@hust.edu.cn}
\cormark[1] 
\cortext[1]{Corresponding author.}

\author[1]{Huan Zhao}[style=chinese]
\ead{huanzhao@hust.edu.cn} 

\author[1]{Zhiao Wei}[style=chinese]
\ead{zhiao_wei@hust.edu.cn} 

\author[1]{Yikun Guo}[style=chinese]
\ead{guoyikun@hust.edu.cn} 

\author[2]{Tiange Wu}[style=chinese]
\ead{wuw_mail@163.com} 

\author[2]{Guoqiang Guo}[style=chinese]
\ead{guoguo_0@163.com} 

\author[1]{Haoyuan Zhou}[style=chinese]
\ead{zhouhaoyuan@hust.edu.cn} 

\author[1]{Jie Pan}[style=chinese]
\ead{panjie@hust-wuxi.com} 

\author[1]{Han Ding}[style=chinese]
\ead{dinghan@hust.edu.cn} 

\affiliation[1]{organization={State Key Laboratory of Intelligent Manufacturing Equipment and Technology, Huazhong University of Science and Technology},
                city={Wuhan},
                postcode={430074}, 
                country={China}}
                
\affiliation[2]{organization={Shanghai Spaceflight Precision Machinery Institute},
                city={Shanghai},
                postcode={201600}, 
                country={China}}

\begin{abstract}
Polishing is a critical finishing process in high-end manufacturing fields such as aerospace, where surface quality directly affects the service performance and reliability of components. Robotic imitation learning provides a flexible solution for such tasks, but current methods remain limited in industrial polishing because of long-horizon dependencies, uncertain stage transitions, and the difficulty of modeling and regulating coupled process parameters. To address these issues, this paper proposes a Stage-Aware and Roughness-Constrained Diffusion Policy (SRDP) for robotic polishing. SRDP infers the process-stage posterior from multimodal observation histories and uses it to condition the shared reverse denoising process, enabling stage-consistent action generation without external stage labels during execution. Furthermore, a roughness-oriented process-constrained diffusion sampling method is incorporated to generate constrained feed speed and normal contact force under stage-wise preset spindle speeds, thereby improving process consistency and physical feasibility. Systematic experiments are conducted on two representative scenarios, namely spacecraft cabin coating-surface polishing and inner-cavity structural surface finishing. Comparisons with advanced baselines, ablation studies, and real-robot validations comprehensively evaluate the proposed method. The results show that SRDP improves stage-transition stability, process-parameter consistency, and final surface quality across different polishing scenarios.
\end{abstract}


\begin{highlights}
\item A unified Stage-Aware and Roughness-Constrained Diffusion Policy (SRDP) framework is proposed to learn multi-stage robotic polishing skills under roughness constraints from demonstrations.

\item Stage posteriors inferred from multimodal histories are introduced into shared reverse denoising, enabling stage-conditioned generation of consistent long-horizon action sequences.

\item Roughness-quality and physical-feasibility constraints guide diffusion sampling to generate feed speed and normal contact force that satisfy process and feasibility requirements under stage-wise preset spindle speeds.

\item Real-robot experiments on spacecraft cabin coating-surface polishing and inner-cavity finishing validate improvements in task performance, parameter stability, and surface quality.
\end{highlights}

\begin{keywords}
  Robotic polishing \sep
  Diffusion policy \sep  
  Imitation learning \sep
  Multi-stage process \sep
  Surface finishing \sep

\end{keywords}

\maketitle

\section{Introduction}
Polishing is an essential finishing process in aerospace and other high-value manufacturing fields, where surface quality directly affects component performance and reliability. However, conventional methods are increasingly limited for small-batch, multi-variety, and complex-surface components \cite{2022_SCTS_Li}. Manual polishing is labor-intensive and inconsistent \cite{2025_TMECH_Ke}, while dedicated machine-tool solutions are often costly and inflexible. Robotic polishing offers a promising alternative through programmability, repeatability, and perception capability \cite{2023_RCIM_Ke}, but many existing systems still rely on offline programming or model-based control, which limits their adaptability to diversified manufacturing scenarios \cite{2020_RCIM_Zhu}.

Robotic imitation learning offers a more flexible and efficient paradigm than traditional offline programming or purely model-based control for the autonomous execution of complex contact-rich tasks. Its key advantage lies in learning policies directly from human demonstrations, without requiring explicit modeling of complex system dynamics or manual design of cost functions. This makes it particularly suitable for manufacturing processes that are difficult to model precisely but can be learned from accumulated demonstrations. Early approaches mainly relied on Gaussian mixture models, task-parameterized movement learning \cite{2016_ISR_Calinon}, and dynamic movement primitives \cite{2006_Springer_Schaal} to represent and reproduce demonstrated trajectories. These methods provide favorable interpretability and trajectory smoothness, but their capability is limited when dealing with high-dimensional visual perception and complex multimodal action distributions. Subsequently, methods such as Generative Adversarial Imitation Learning (GAIL) formulated imitation learning as adversarial policy learning \cite{2016_NeurIPS_Ho}, Behavior Transformer (BeT) employed Transformers to model multimodal action distributions \cite{2022_NeurIPS_Shafiullah}, and Action Chunking with Transformers (ACT) introduced chunked action prediction to mitigate error accumulation in step-wise behavior cloning \cite{2023_arXiv_Zhao}. In recent years, with the development of large-scale robot learning and generative policies, methods such as RT-1 \cite{2022_arXiv_Brohan}, RT-2 \cite{2023_CoRL_Zitkovich}, and Diffusion Policy \cite{2025_IJRR_Chi} have further demonstrated the strong potential of end-to-end policies in multi-task generalization, high-dimensional continuous action modeling, and temporal consistency preservation. In particular, Diffusion Policy represents robotic policies as a conditional denoising diffusion process, showing significant advantages in continuous action generation and multimodal policy representation. It has achieved promising results in assembly\cite{2023_NeurIPS_Giannone_a}, grasping\cite{2024_arXiv_Ze}, and deformable object manipulation tasks \cite{2024_RAL_Scheikl_a}, thereby providing a new technical foundation for action generation in complex manufacturing processes.

Despite the significant progress of diffusion-based imitation learning in robotic manipulation, directly applying it to polishing and related manufacturing processes remains challenging. First, these processes exhibit long-horizon process-stage uncertainty, where visually similar local states may correspond to different process stages and control objectives. Second, diffusion-based policies generate actions through stochastic denoising, and the native randomness of this process may cause unconstrained process-parameter drift when feed speed and normal contact force are generated under stage-wise preset spindle-speed settings. Since these parameters jointly determine material removal and surface roughness evolution \cite{2023_RCIM_Wang,2021_JMP_Li}, such drift can lead to physically inconsistent parameter combinations, unstable contact, surface-quality fluctuations, or local over-polishing. To the best of our knowledge, there is still no unified solution that jointly handles long-horizon process-stage uncertainty and unconstrained process-parameter drift in high-dimensional generative policies for robotic polishing.

The main goal of this article is to develop a physically credible behavior-learning and generation framework for complex manufacturing processes. To this end, we propose a stage-aware and roughness-constrained diffusion policy (SRDP), which combines stage-conditioned reverse denoising and roughness-constrained diffusion sampling to address long-horizon process-stage uncertainty and unconstrained process-parameter drift. To the best of our knowledge, this is the first attempt to introduce such a unified stage-aware and quality-constrained diffusion formulation into multi-stage robotic polishing. Beyond the specific choice of polishing parameters, the same formulation can serve as a general mechanism for incorporating stage-dependent objectives, physical constraints, and quality-driven regulation into generative robotic policies, making it applicable to a broader class of manufacturing processes such as grinding, drilling, deburring, and precision finishing.

To clarify our motivation, we further review studies on long-horizon process-stage uncertainty and unconstrained process-parameter drift before presenting our method and contributions.

\begin{figure*}
    \centering
    \includegraphics[width=1.0\textwidth]{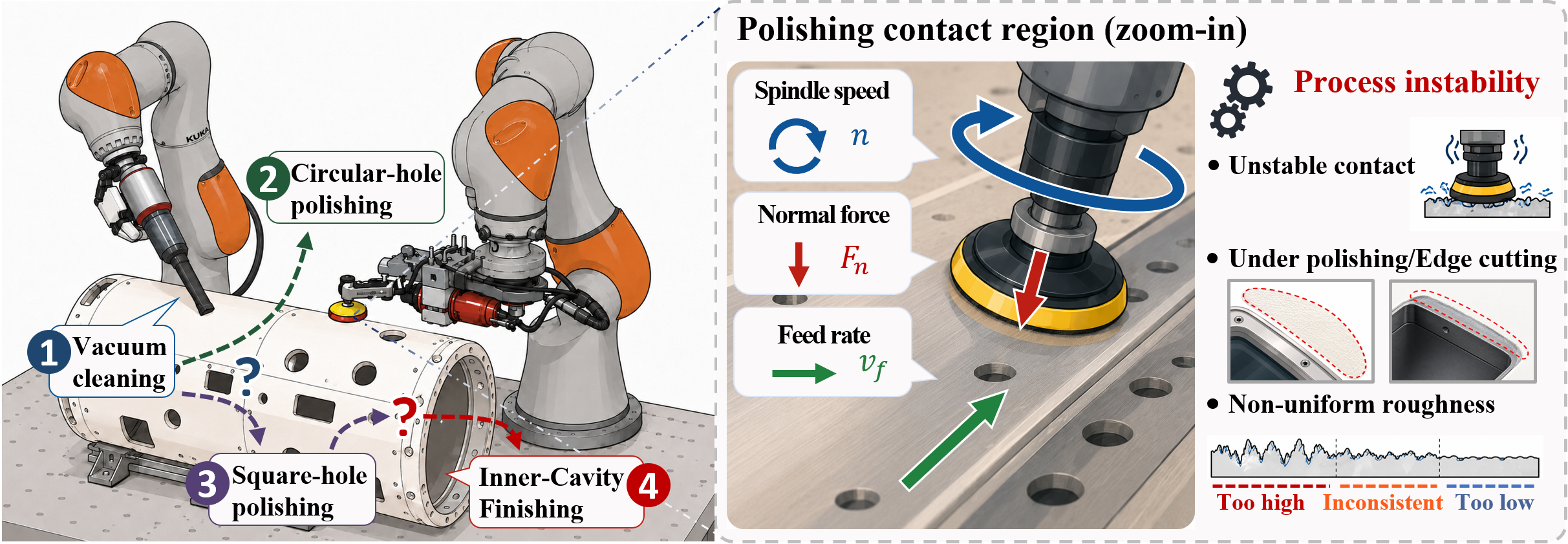}
    \caption{Illustration of the two main challenges in multi-stage robotic polishing of spacecraft cabin sections. Although shown in a hand-drawn style, the figure is redrawn from the real experimental platform and representative polishing results. The first challenge is uncertain next-stage decision making in long-horizon workflows, where the robot must select the correct subsequent operation among similar local regions and multiple process branches. The second challenge is the coordination of feed speed and normal contact force under stage-wise preset spindle speeds, whose mismatch may cause process instability and surface defects such as under-polishing, edge cutting, and non-uniform roughness.}
    \label{fig:motivation}
\end{figure*}

\subsection{Long-Horizon Process-Stage Uncertainty}

Existing studies on multi-stage long-horizon robotic manipulation have mainly evolved along three lines: explicit task organization, structured skill modeling, and high-level condition-driven long-horizon policy learning. Explicit task decomposition methods improve the stability and recovery capability of long-horizon task execution by separating high-level task organization from low-level motion execution. The hierarchical imitation learning framework proposed by Luo et al. \cite{2024_TRO_Luo} represents a class of methods that enhance the execution stability of long-horizon tasks through explicit stage organization. This indicates that the key to multi-stage tasks lies not only in improving single-step control accuracy, but also in maintaining the action logic and recovery mechanism across different stages. Similarly, the mixed reality-assisted human-to-robot skill transfer approach proposed by Wu et al. \cite{2026_RCIM_Wu} organizes complex assembly procedures into a set of transferable skill units through mixed reality demonstration, generative visuomotor imitation learning, and contact-rich primitives. This further suggests that, in industrial tasks, intermediate skill representations themselves are an important support for stable long-horizon execution. Another line of work focuses on the structured representation and composition mechanism of skill units. For example, Learning a Skill-sequence-dependent Policy \cite{2021_CASE_Li} and LeSkill \cite{2025_TSMCS_Huang} enhance skill composition in long-horizon tasks from the perspectives of skill-order dependency and reusable skill libraries, respectively. Logic-LfD \cite{2024_RAL_Zhang} and long-horizon planning frameworks based on behavior trees or Planning Domain Definition Language (PDDL) \cite{2025_IROS_Chen} further integrate skill sequences with task logic and reactive replanning, enabling the system not only to organize multi-step procedures but also to maintain task structure in dynamic environments. Different from the above studies, FIT \cite{2025_TCDS_Salam}, Diffusion Trajectory-Guided Policy \cite{2025_RAL_Fan}, and Hierarchical Diffusion Policy \cite{2024_CVPR_Ma} place more emphasis on modeling long-horizon dependencies through future-state prediction, high-level trajectory guidance, and hierarchical conditional generation, thereby advancing long-horizon policy learning from explicit task decomposition toward high-level condition-driven generative control. Overall, existing studies have significantly improved task organization and cross-stage planning in long-horizon tasks. However, their high-level conditions are still mainly represented as skill units, subgoals, future trajectories, or planning prompts, rather than being directly oriented toward the problem of posterior stage inference in complex manufacturing processes, where the stage is a latent process state inferred from multimodal histories rather than a readily available command label. For multi-process tasks such as polishing, different stages often exhibit highly similar local observations, while their action patterns, contact states, and control objectives can be substantially different. Therefore, task organization and high-level planning alone are insufficient to support stable process-stage transitions. A policy mechanism is further required to infer the current stage from multimodal observation histories and to directly drive robot action generation through stage-conditioned control.

\subsection{Unconstrained Process-Parameter Drift}

Existing studies have also explored how to incorporate geometric priors, physical constraints, and process knowledge into robotic action generation. Earlier works mainly focused on enhancing perception and action representations. 3D Diffusion Policy \cite{2024_arXiv_Ze}, Diffusion-EDFs \cite{2024_CVPR_Ryu}, and Movement Primitive Diffusion \cite{2024_RAL_Scheikl_b} improve the geometric accuracy, structural consistency, and trajectory smoothness of action generation from the perspectives of 3D geometric representations, SE(3) symmetry, and movement-primitive parameter spaces, respectively. These methods improve the representation of action intent and can produce higher-quality outputs through better parameterization, but they still have difficulty in directly ensuring that the generated results satisfy specific process objectives. Another line of research attempts to establish the connection between skill learning and machining outcomes through process-mechanism models. AL-ProMP and related material removal rate models \cite{2023_RCIM_Wang,2021_JMP_Li} have linked force-relevant skills with polishing outcomes and material removal behavior, indicating that the coupled relationship among contact force, feed speed, and tool action determines the final surface quality. However, these methods are mostly built upon traditional explicit skill learning or process modeling. Although they are advantageous in force-relevant skill generalization and process-outcome interpretation, they are less capable of fully exploiting high-dimensional visual information and are also less flexible than generative policies in handling multimodal action distributions in complex tasks. Building on these studies, more recent works have begun to introduce constraints directly into the generation or execution process. Physics-Informed Diffusion Models \cite{2025_ICLR_Bastek} and Diffusion Optimization Models \cite{2023_NeurIPS_Giannone_b} inject constraints into the sampling process through physical-equation losses and trajectory-alignment mechanisms, respectively. Diffusion Policies for Dynamically Admissible Trajectories (DDAT) \cite{2025_arXiv_Bouvier} introduces dynamic reachable-set projection during both training and inference, enabling diffusion trajectories to satisfy dynamic feasibility constraints. Adaptive Compliance Policy and the tactile-diffusion variable-impedance control framework \cite{2025_ICRA_Hou,2025_RAL_Li} further incorporate contact compliance into policy generation, allowing position control and contact force regulation to work in a coordinated manner. Compared with the first two categories of methods, these studies have advanced constraints from representation learning and process analysis to the action generation process. However, their targets are still mainly general physical consistency, dynamic feasibility, trajectory executability, or contact compliance, rather than manufacturing quality objectives such as surface roughness. For robotic polishing and grinding, the final surface quality depends not only on whether an action is executable, but also on whether contact force, end-effector compliance, and the coupled process parameters including spindle speed, feed speed, and normal contact force are handled in a unified manner \cite{2019_RCIM_Chen,2022_RCIM_Wei,2021_JMP_Wang,2023_JMP_Mu}. Therefore, although existing studies have advanced constrained generation and process modeling along separate routes, there is still a lack of a method that can directly introduce roughness-driven joint parameter constraints into high-dimensional generative policies, so as to generate and regulate actions for real manufacturing processes.

Overall, existing studies have improved long-horizon task organization and constrained action generation from different perspectives. However, existing methods remain insufficient for robotic polishing tasks, where process-stage uncertainty and process-parameter drift coexist. Therefore, this work aims to develop a diffusion-based learning framework to address these challenges.

\subsection{Contributions of This Article}

To address the above issues, this paper develops SRDP as a unified diffusion policy learning framework for robotic polishing tasks. First, a stage inference network is constructed from multimodal observation histories, and stage-conditioned reverse denoising is introduced to generate stage-consistent action sequences under long-horizon process-stage uncertainty. Then, based on a roughness-oriented process model, roughness-consistency and physical-feasibility constraints are incorporated into constraint-guided diffusion sampling to generate constrained feed speed and normal contact force under stage-wise preset spindle speeds. Finally, the effectiveness of the proposed method is validated in coating-surface polishing and inner-cavity finishing experiments. The main contributions of this paper are summarized as follows.

\begin{enumerate}
    \item SRDP is proposed for multi-stage robotic polishing to learn polishing skills from demonstrations; it infers stage posteriors from multimodal histories and conditions shared reverse denoising for stage-consistent action generation.

    \item A process-constrained diffusion sampling method is proposed to introduce roughness-quality and physical-feasibility constraints into the reverse denoising process, generating constrained feed speed and normal contact force under stage-wise preset spindle speeds.

    \item Extensive experiments, including comparisons with advanced methods, module ablation studies, and real-world scenario validations, demonstrate the advantages of the proposed method in terms of task performance, parameter stability, and surface quality.
\end{enumerate}

The remainder of this paper is organized as follows. Section II presents the problem decomposition and overall framework. Section III introduces the stage inference network based on multimodal observation histories and the stage-conditioned reverse denoising mechanism. Section IV describes the roughness-oriented process modeling and the process-constrained diffusion sampling method. Section V presents the experimental setup, comparative results, and ablation analysis. Section VI concludes this paper and discusses future work.

\section{Framework}
This section introduces the diffusion policy learning framework proposed for complex robotic polishing tasks, aiming to develop a robotic polishing action generation method that can simultaneously model task-stage information and the coordinated relationships among process parameters. Specifically, Section 2-A formulates the problem decomposition for complex polishing tasks, and Section 2-B presents the overall pipeline and basic components of the proposed framework.

\subsection{Problem Decomposition}
Complex robotic polishing tasks differ fundamentally from conventional single-stage trajectory tracking problems. In essence, they are action generation problems that simultaneously involve multi-stage temporal organization and coordinated decision-making over multiple process parameters. In the task scenarios considered in this work, namely spacecraft cabin coating-surface polishing and inner-cavity structural polishing, the complete workflow is not represented by a single continuous trajectory, but rather consists of multiple stages organized according to the process sequence, such as workpiece placement, edge polishing, surface polishing, assembly-surface polishing, chamfer polishing, and vacuum cleaning. These stages differ not only in visual appearance, but more importantly in contact conditions, motion patterns, and control objectives. For instance, polishing stages emphasize uniform surface finishing, stable normal contact, and consistent trajectory coverage, whereas vacuum cleaning and workpiece handling are auxiliary operations without material removal, whose action characteristics, contact modes, and task objectives differ significantly from those of polishing. Therefore, for such complex tasks, the robot must generate stage-consistent actions according to the current process phase, rather than treating the entire task as a direct one-step mapping from observations to actions.

Meanwhile, the polishing process also exhibits a pronounced multi-parameter coupling characteristic. The final machining quality is determined not only by the geometric trajectory of the end-effector, but also by process parameters such as tool rotational speed, feed speed, and normal contact force. More importantly, these parameters are not independent of each other; instead, they jointly affect the material removal process and the evolution of the final surface quality. For example, under the same geometric trajectory, different combinations of rotational speed, feed speed, and normal force may correspond to entirely different contact conditions and roughness outcomes. Therefore, complex polishing tasks should not be regarded as purely geometric motion generation problems, but rather as joint decision-making problems that simultaneously involve geometric action generation and coordinated process-parameter planning.

\begin{figure*}[!t]
\centering
\includegraphics[width=0.95\textwidth]{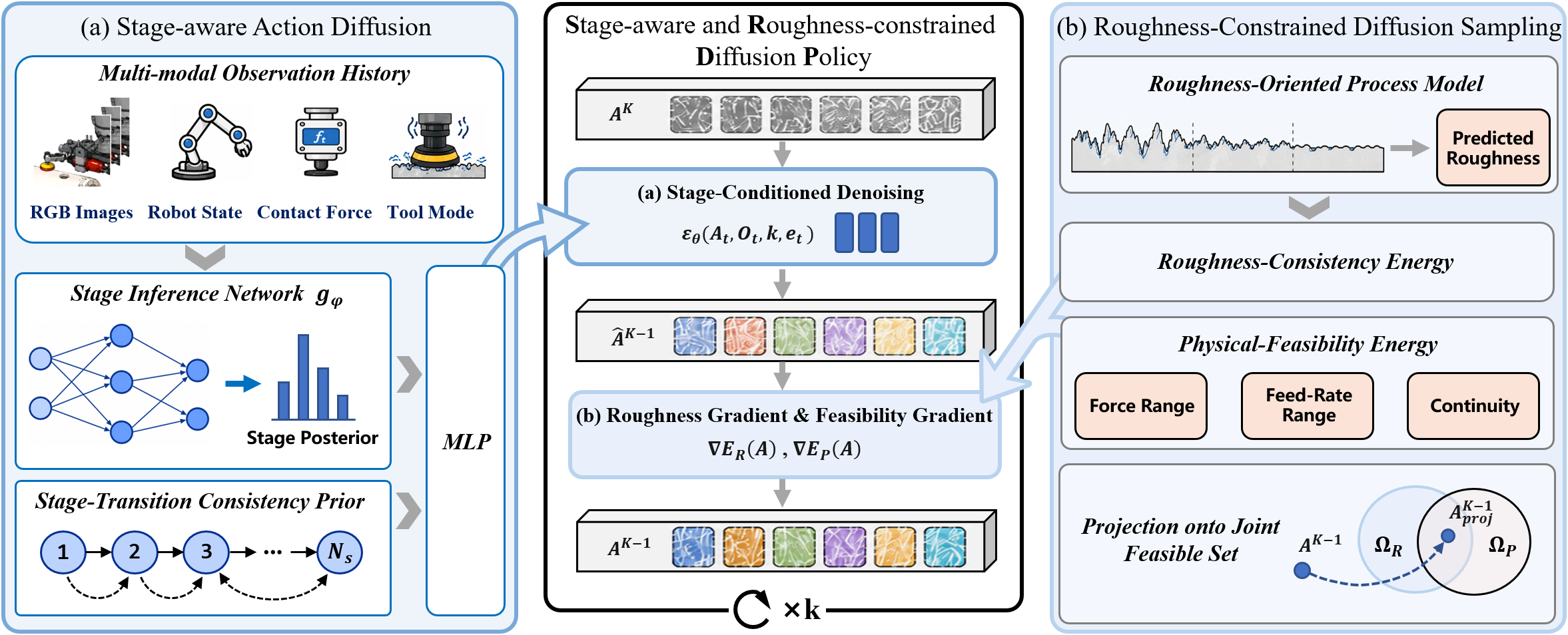}
\caption{Overview of the proposed SRDP framework for multi-stage robotic polishing. The left branch extracts stage information from multimodal observation histories for stage-aware action diffusion (Section~3), the center block performs stage-conditioned denoising within the action diffusion process (Section~3), and the right branch imposes roughness-consistency and physical-feasibility guidance on process parameters during sampling through roughness-oriented process modeling (Section~4.1) and constraint-guided diffusion sampling (Section~4.2). The framework finally outputs stage-consistent and process-feasible polishing actions.}
\label{fig:framework}
\end{figure*}

Based on the above analysis, the complete task is represented as a stage set consisting of $N_s$ stages:
\begin{equation}
\mathcal{S}=\{s_1,s_2,\ldots,s_{N_s}\},
\label{eq:stage_set}
\end{equation}
where $N_s$ denotes the number of process stages, and $s_i \in \{1,\ldots,N_s\}$ denotes the $i$-th task stage. Each stage is introduced to characterize the differences in action distributions and process patterns across distinct stages.

At discrete time step $t$, the robot receives an observation history window of length $L$:
\begin{equation}
\mathbf{O}_t=\{o_{t-L+1},o_{t-L+2},\ldots,o_t\},
\label{eq:obs_history}
\end{equation}
where each single-step observation $o_t$ is composed of multimodal information and can be uniformly written as
\begin{equation}
o_t=\{I_t,x_t,f_t,m_t\},
\label{eq:single_obs}
\end{equation}
where $I_t$ denotes the visual observation, $x_t$ denotes the robot state and end-effector pose, $f_t$ denotes the normal contact force, and $m_t$ denotes the tool mode. The use of an observation history, rather than a single-frame observation, is motivated by the fact that stage recognition and action decision-making in polishing depend not only on the current image, but also on recent motion evolution and changes in contact conditions.

Given the current observation history, the policy is required to predict a future action sequence of horizon $H$, denoted by $\mathbf{A}_t=\{a_t,a_{t+1},\ldots,a_{t+H-1}\}$, where each action is written as $a_t=[p_t,r_t,\omega_t,v_t,F_t^n,u_t]$. Here, $p_t \in \mathbb{R}^6$ denotes the dual-arm end-effector position command, $r_t \in \mathbb{R}^6$ denotes the dual-arm orientation-vector command, $\omega_t$ denotes the stage-wise preset tool rotational speed assigned according to the current process stage, $v_t \in \mathbb{R}$ denotes the feed speed, $F_t^n \in \mathbb{R}$ denotes the normal contact force, and $u_t$ denotes auxiliary tool-mode commands, such as spindle on/off or vacuum gripper grasp/release commands, which are generated as continuous outputs and discretized by thresholding before execution.

Accordingly, the complex polishing task considered in this paper can be formulated as a conditional action-sequence generation problem, where the future action sequence $\mathbf{A}_t=\{a_t,a_{t+1},\ldots,a_{t+H-1}\}$ is predicted conditioned on the observation history $\mathbf{O}_t=\{o_{t-L+1},o_{t-L+2},\ldots,o_t\}$. Under this formulation, the policy model must simultaneously address two key issues: 1) how to generate action patterns that are consistent with the current task stage, and 2) how to generate feed speed and normal contact force under stage-wise preset tool rotational speeds.

\subsection{Pipeline for SRDP}
Based on the above problem formulation, the proposed SRDP is formulated as a unified SRDP generation process, as shown in Fig.~\ref{fig:framework}. The figure should be read from left to right: multimodal observation histories first provide process-stage information, the central diffusion policy generates action sequences through iterative reverse denoising, and the roughness-constrained sampling branch regulates process parameters during reverse denoising. Therefore, the pipeline is not an additional module, but a unified formulation that connects the problem decomposition in Section~2 with the two technical components developed in Sections~3 and 4.

Specifically, the left part of Fig.~\ref{fig:framework} corresponds to the stage-aware action diffusion component. During online execution, the robot receives the current multimodal observation history $\mathbf{O}_t=\{o_{t-L+1},o_{t-L+2},\ldots,o_t\}$, including RGB images, robot states, contact force, and tool-mode information. A stage inference network estimates the posterior distribution of the current process stage, while a stage-transition consistency prior improves the temporal stability of the inferred stage sequence. The resulting stage representation is then injected into the central reverse denoising process, enabling the action generator to produce a candidate future action sequence $\mathbf{A}_t=\{a_t,a_{t+1},\ldots,a_{t+H-1}\}$ that is consistent with the current process stage. This part of the formulation is developed in Section~3.

The right part of Fig.~\ref{fig:framework} corresponds to the roughness-constrained diffusion sampling component. After each denoising step produces candidate actions, the stage-wise preset spindle speed and the generated feed speed and normal contact force are evaluated using a roughness-oriented process model. The resulting roughness-consistency energy and physical-feasibility energy provide guidance gradients for diffusion sampling, and the final process parameters are projected onto a joint feasible set. This part of the formulation is developed in Section~4, where the roughness model serves as a process-quality constraint rather than an independent contribution. Through this left-to-right information flow, SRDP embeds stage-conditioned action generation and quality-feasibility parameter regulation into the same diffusion sampling process.

\section{Stage-Conditioned Action Diffusion Policy}
This section focuses on stage-conditioned action sequence generation for complex robotic polishing tasks. The key issue is how to generate fine-grained action sequences that are consistent with the current process stage and its contextual information, while maintaining smooth and feasible transitions across adjacent stages. To this end, a stage-conditioned action diffusion policy is constructed by incorporating stage inference, stage-specific denoising, and transition-consistency regularization into the reverse denoising process. Unlike hierarchical imitation learning with hard policy selection, the inferred stage posterior is used here as a conditioning variable of the shared denoising network. This enables the policy to generate stage-consistent action sequences for multi-stage polishing tasks within the diffusion sampling framework.

\subsection{Prior Knowledge of Diffusion Policy}
Diffusion Policy (DP) formulates visuomotor action generation as a conditional denoising diffusion process. Let $\mathbf{A}_t^{0}$ denote an expert action sequence of horizon $H$ starting from control step $t$. Following the standard denoising diffusion formulation, Gaussian noise is gradually injected into the action sequence over $K$ diffusion steps, with $\alpha_k$ denoting the noise scheduling parameter and $\bar{\alpha}_{k}=\prod_{i=1}^{k}\alpha_i$. The noisy action sequence at diffusion step $k$ can then be written as
\begin{equation}
\mathbf{A}_t^{k}=\sqrt{\bar{\alpha}_{k}}\,\mathbf{A}_t^{0}+\sqrt{1-\bar{\alpha}_{k}}\,\epsilon,\qquad \epsilon \sim \mathcal{N}\left( 0,\mathbf{I} \right).
\label{eq:noisy_action_seq}
\end{equation}

The goal of the reverse process is to progressively recover the clean action sequence $\mathbf{A}_t^{0}$ from $\mathbf{A}_t^{K}$ conditioned on the observation history. A conventional diffusion policy employs a denoising network $\epsilon_{\theta}\left( \mathbf{A}_t^{k},\mathbf{O}_{t},k \right)$ to predict the injected noise, and its reverse update takes the form
\begin{equation}
\mathbf{A}_t^{k-1}
=
\frac{1}{\sqrt{\alpha_{k}}}
\left(
\mathbf{A}_t^{k}
-
\frac{1-\alpha_{k}}{\sqrt{1-\bar{\alpha}_{k}}}
\epsilon_{\theta}\left( \mathbf{A}_t^{k},\mathbf{O}_{t},k \right)
\right)
+
\sigma_{k}\xi,
\label{eq:reverse_update_basic}
\end{equation}
where $\xi \sim \mathcal{N}\left( 0,\mathbf{I} \right)$ and $\sigma_{k}$ denotes the weight of the stochastic term at the $k$-th sampling step.

\subsection{Stage Inference and Stage-Conditioned Reverse Denoising}

Reliable stage inference cannot rely only on a single-frame image. Many polishing stages may exhibit similar visual appearances at a certain instant, especially when the tool remains close to the workpiece throughout the task. In contrast, human experts usually infer the current process stage by jointly considering scene semantics, motion evolution, and contact-state changes. Based on this observation, this paper uses the multimodal observation history $\mathbf{O}_t$ defined in Section~II-A as the input to the stage inference module, where each observation consists of the visual observation, robot state and end-effector pose, normal contact force, and tool mode. Compared with a single-frame observation, the observation history provides temporal cues for distinguishing process stages with similar instantaneous visual appearances.

We define a stage inference network $g_{\phi}$, which maps the multimodal observation history $\mathbf{O}_t$ to the posterior probability distribution over the current process stage:
\begin{equation}
    \mathbf{z}_{t}
    =
    g_{\phi}\left(\mathbf{O}_{t}\right),
    \qquad
    \mathbf{z}_{t}\in \Delta^{N_s-1},
    \label{eq:stage_posterior}
\end{equation}

where $\mathbf{z}_{t}$ denotes the stage posterior over $N_s$ process stages:
\begin{equation}
    \mathbf{z}_{t}
    =
    [z_{t}^{(1)},\ldots,z_{t}^{(N_s)}],
    \qquad
    \sum_{j=1}^{N_s} z_{t}^{(j)}=1.
    \label{eq:stage_prob_sum}
\end{equation}
Here, $z_{t}^{(j)}$ represents the posterior probability that the current observation history belongs to the $j$-th process stage.

To condition the reverse denoising process on the inferred process stage, the stage posterior is first encoded into a stage condition vector as
\begin{equation}
    \mathbf{e}_{t}
    =
    \mathrm{MLP}_{s}
    \left(
    \mathbf{z}_{t}
    \right),
    \label{eq:stage_embedding}
\end{equation}
where $\mathrm{MLP}_{s}(\cdot)$ denotes the multilayer perceptron (MLP)-based stage embedding network. The stage condition vector $\mathbf{e}_{t}$ is then used as an additional conditioning input to the shared denoising network. The noise prediction is written as
\begin{equation}
    \hat{\epsilon}
    =
    \epsilon_{\theta}
    \left(
    \mathbf{A}_t^{k},
    \mathbf{O}_{t},
    k,
    \mathbf{e}_{t}
    \right).
    \label{eq:stage_conditioned_noise}
\end{equation}
Here, $\epsilon_{\theta}$ denotes the denoising network shared by all process stages, and $\hat{\epsilon}$ denotes the predicted noise. In this way, the stage posterior conditions the shared denoising network through $\mathbf{e}_{t}$, enabling the reverse diffusion process to generate action sequences that are consistent with the current process context.

After obtaining the stage-conditioned noise prediction, the reverse diffusion update can be written as
\begin{equation}
    \mathbf{A}_t^{k-1}
    =
    \frac{1}{\sqrt{\alpha_{k}}}
    \left(
    \mathbf{A}_t^{k}
    -
    \frac{1-\alpha_{k}}{\sqrt{1-\bar{\alpha}_{k}}}
    \hat{\epsilon}
    \right)
    +
    \sigma_{k}\boldsymbol{\xi},
    \label{eq:stage_conditioned_reverse_update}
\end{equation}
where $\boldsymbol{\xi}\sim\mathcal{N}(\mathbf{0},\mathbf{I})$, and $\sigma_{k}$ denotes the weight of the stochastic term at the $k$-th sampling step. During each denoising step, the action sequence is not recovered solely from the observation-conditioned demonstration distribution. Instead, the denoising process is further guided by the stage condition associated with the current process context, thereby enhancing stage-semantics preservation in long-horizon multi-process tasks.

For industrial polishing tasks, stage transitions do not occur arbitrarily, but usually follow an approximately ordered process flow. For example, vacuum cleaning usually occurs after polishing rather than before workpiece placement, and chamfering is also unlikely to occur before assembly-surface handling. Since the stage posterior is used to condition the reverse denoising process, its temporal stability is important for smooth action generation across adjacent stages. Therefore, this paper further introduces a stage-transition consistency prior to improve the temporal stability of stage inference.

We define the stage-transition matrix $\mathbf{T}\in\mathbb{R}^{N_s\times N_s}$ as
\begin{equation}
    T_{ij}
    =
    p\left(s_t=j \mid s_{t-1}=i\right).
    \label{eq:transition_matrix}
\end{equation}
With this definition, $\mathbf{T}$ is row-stochastic, i.e., $\sum_{j=1}^{N_s}T_{ij}=1$, and $\mathbf{T}^{\top}\mathbf{z}_{t-1}$ gives the stage prior predicted from the previous posterior. According to the specific task, $\mathbf{T}$ can be initialized based on prior knowledge of the process flow, or estimated from labeled demonstration data. To enhance the consistency of stage posteriors between adjacent time steps, the transition-consistency regularization is written as a soft-label cross-entropy loss:
\begin{equation}
    \mathcal{L}_{\mathrm{trans}}
    =
    -
    \sum_{t}
    \left(\mathbf{T}^{\top}\mathbf{z}_{t-1}\right)^{\top}
    \log \mathbf{z}_{t},
    \label{eq:transition_loss}
\end{equation}
where the logarithm is applied element-wise. This term directly uses the stage prior propagated from the previous posterior to supervise the current stage posterior, thereby penalizing abrupt stage changes that are inconsistent with the process flow and reducing oscillatory predictions near stage boundaries.

\subsection{Training Objective and Receding-Horizon Execution}

The overall training objective of the proposed framework consists of the diffusion denoising loss, the stage-posterior supervision loss, and the transition-consistency regularization:
\begin{equation}
\mathcal{L}
=
\mathcal{L}_{\mathrm{diff}}
+
\lambda_s \mathcal{L}_{\mathrm{stage}}
+
\lambda_t \mathcal{L}_{\mathrm{trans}}.
\label{eq:total_loss}
\end{equation}

The diffusion denoising loss is defined as
\begin{equation}
\mathcal{L}_{\mathrm{diff}}
=
\mathbb{E}_{\mathbf{A}_t^{0},\epsilon,k}
\left[
\left\|
\epsilon-\hat{\epsilon}
\right\|_{2}^{2}
\right],
\label{eq:diff_loss}
\end{equation}
where the noisy action sequence $\mathbf{A}_t^{k}$ is obtained by adding Gaussian noise to the clean action sequence $\mathbf{A}_t^{0}$ at the $k$-th diffusion step, and $\hat{\epsilon}$ denotes the stage-conditioned noise prediction.

Since the demonstration data are annotated with stage labels $y_t$, the stage inference supervision loss is defined in the form of cross-entropy:
\begin{equation}
\mathcal{L}_{\mathrm{stage}}
=
-
\sum_{t}
\sum_{j=1}^{N_s}
\mathbb{I}\left(y_t=j\right)
\log z_t^{(j)}.
\label{eq:stage_loss}
\end{equation}
In this work, the stage labels are directly obtained from the process workflow annotations. In more general cases, this supervision term can also be approximately constructed using weak labels generated from task scripts, execution logs, or event-switching rules. These labels are used only for training stage inference; during execution, the stage posterior is inferred online from $\mathbf{O}_t$ and used to condition the shared denoising network.

After the reverse diffusion process generates the predicted action sequence $\hat{\mathbf{A}}^{0}$, the system does not execute the entire sequence in a fully open-loop manner. Instead, a receding-horizon execution strategy is adopted: only the first $H_e$ actions are executed at each control cycle, where $H_e < H$. The system then acquires new observations and performs the next round of action generation.

\section{Roughness-Constrained Diffusion Sampling}
In this work, physical and roughness-quality constraints are embedded directly into the reverse diffusion sampling process to regulate the generated process parameters. We first develop a roughness-oriented process modeling method to establish the relationship between process parameters and surface quality, and define a quality-feasible region. We then present a quality-constrained diffusion sampling method, which projects the generated process parameters into the feasible region during denoising, thereby ensuring that the final polishing actions are both physically executable and compliant with the surface-quality requirements.

\subsection{Roughness-Oriented Process Modeling}

In robotic polishing, the surface roughness $R_a$ is one of the most important quality evaluation metrics. Existing studies have shown that roughness is jointly affected by multiple process parameters, including the normal force $F_t^n$, the tool surface linear speed, and the feed speed, while also being related to geometric factors such as the local surface curvature. Referring to the empirical modeling practice for the relationship between surface roughness and process parameters in existing studies \cite{2014_IJAMT_Zhao,2002_IJAMT_Feng}, and considering that the logarithmically transformed linear-regression form provides good interpretability of model parameters in empirical roughness modeling, the local surface roughness prediction is formulated in the following power-law form:
\begin{equation}
\hat{R}_{a,t}=C_{s_t}(F_t^n)^{\alpha}(\omega_t)^{\beta}(v_t)^{\gamma}
\label{eq:roughness_model}
\end{equation}
where $\hat{R}_{a,t}$ denotes the predicted roughness at control step $t$; $C_{s_t}$ is the stage-dependent empirical coefficient; $F_t^n$ is the normal contact force; $\omega_t$ is the spindle angular speed; $v_t$ is the feed speed; and $\alpha$, $\beta$, and $\gamma$ are empirical exponents. It should be emphasized that the training observations do not contain roughness measurements. Therefore, roughness is not used as an input to the diffusion policy. Instead, the network still receives visual, state, force, and stage information, while during inference the stage-wise preset spindle speed and the candidate feed-force parameters are substituted into Eq.~\eqref{eq:roughness_model} to estimate the roughness for subsequent constraint regulation.

Taking the logarithm of Eq.~\eqref{eq:roughness_model} yields
\begin{equation}
\log \hat{R}_{a,t}=\log C_{s_t}+\alpha \log F_t^n+\beta \log \omega_t+\gamma \log v_t
\label{eq:log_roughness_model}
\end{equation}
Equation~\eqref{eq:log_roughness_model} indicates that the model can be calibrated by linear regression using offline experimental data, thereby obtaining the stage-dependent empirical coefficients and sensitivity exponents. In this work, $\alpha$, $\beta$, $\gamma$, and $C_s$ are not manually selected hyperparameters, but are identified offline from calibration experiments and fixed during policy inference for roughness prediction and constraint guidance. This logarithmic linearization is commonly adopted in empirical surface-roughness modeling [3].

For a rotary polishing tool, the spindle speed determines the tool surface linear speed through the effective tool radius, whereas the feed speed mainly affects the local dwell time along the path. Therefore, under a fixed spindle speed, a lower feed speed leads to a longer dwell time over the local region. In many practical polishing stages, the spindle speed is usually pre-specified according to process experience. Consistent with the experimental platform used in this work, SRDP does not generate spindle speed at every control step; instead, spindle speed is preset according to the current process stage. For a given stage $s_t$, the spindle speed can therefore be regarded as a constant:
\begin{equation}
\omega_t=\bar{\omega}^{(s_t)},\qquad t\in \mathrm{stage}\ s_t
\label{eq:constant_spindle}
\end{equation}
Substituting Eq.~\eqref{eq:constant_spindle} into Eq.~\eqref{eq:roughness_model}, the stage-wise roughness prediction model can be further simplified as
\begin{equation}
\hat{R}_{a,t}=\bar{C}_{s_t}(F_t^n)^{\alpha}(v_t)^{\gamma}
\label{eq:stagewise_roughness_model}
\end{equation}
where $\bar{C}_{s_t}=C_{s_t}(\bar{\omega}^{(s_t)})^{\beta}$. This implies that, under the assumption of stage-wise fixed spindle speed, the diffusion policy mainly needs to generate and constrain the normal force $F_t^n$ and the feed speed $v_t$, and these two variables directly determine whether the target roughness requirement can be satisfied.

Based on this relationship, the target roughness $R_a^\star$, the allowable roughness error $\varepsilon_R$, and the executable ranges of the normal force and feed speed jointly determine whether a candidate parameter pair is quality-feasible. The formal roughness-consistency and physical-feasibility sets used for diffusion sampling are defined in Section~4.2.

In addition, to ensure physical realizability at the robot execution level, the process parameters must also satisfy temporal continuity constraints:
\begin{equation}
|F_t^n-F_{t-1}^n| \le r_F\Delta t
\label{eq:force_continuity}
\end{equation}
\begin{equation}
|v_t-v_{t-1}| \le r_v\Delta t
\label{eq:feed_continuity}
\end{equation}
where $r_F$ and $r_v$ denote the maximum variation rates of the normal force and feed speed, respectively, and $\Delta t$ is the control period. Equations~\eqref{eq:force_continuity} and \eqref{eq:feed_continuity} ensure temporal continuity of the process parameters and prevent abrupt parameter changes that are difficult for the robot to track.

\subsection{Constraint-Guided Diffusion Sampling}
In this work, the roughness prediction model is treated as a joint constraint on the generation of process parameters during diffusion, so as to suppress the disorderly drift of process parameters in reverse denoising and ensure that the generated normal contact force $F_t^n$ and feed speed $v_t$ satisfy the target roughness requirement under the stage-wise preset spindle speed $\bar{\omega}^{(s_t)}$, thereby improving the controllability and consistency of the final surface roughness.

For a given time step $t$, let the target roughness be denoted by $R_{a,t}^{\star}$. According to the roughness model in Section~IV-A, the set of process parameters satisfying this target roughness is not a collection of isolated points, but rather a coupled relation induced by the roughness equation in the parameter space. When the spindle speed is fixed within stage $s_t$ as $\bar{\omega}^{(s_t)}$, the roughness prediction model can be written as
\begin{equation}
\hat{R}_{a,t}=\bar{C}_{s_t}(F_t^n)^{\alpha}(v_t)^{\gamma}
\label{eq:roughness_stagewise_again}
\end{equation}
Accordingly, the parameter-consistency relation corresponding to the target roughness $R_{a,t}^{\star}$ can be expressed as
\begin{equation}
\hat{R}_{a,t}(F_t^n,v_t)=R_{a,t}^{\star}.
\label{eq:roughness_target_relation}
\end{equation}

Considering model errors, process disturbances, and sensing noise, exact equality is generally not required in practice. Instead, the parameter pair is allowed to lie within a tolerance band around the target roughness. To this end, the roughness-consistency set is defined as
\begin{equation}
\Omega_{R,t}
=
\left\{
(F_t^n,v_t)
\ \middle|\
\left|
\log \hat{R}_{a,t}-\log R_{a,t}^{\star}
\right|
\le
\varepsilon_R
\right\}
\label{eq:roughness_consistency_set}
\end{equation}
Substituting Eq.~\eqref{eq:roughness_stagewise_again} into Eq.~\eqref{eq:roughness_consistency_set} yields
\begin{equation}
\Omega_{R,t}
=
\left\{
(F_t^n,v_t)
\ \middle|\
\left|
\alpha \log F_t^n+\gamma \log v_t-b_t
\right|
\le
\varepsilon_R
\right\}
\label{eq:roughness_consistency_set_expanded}
\end{equation}
with
\begin{equation}
b_t=\log R_{a,t}^{\star}-\log \bar{C}_{s_t}.
\label{eq:b_t_def}
\end{equation}
Equation~\eqref{eq:roughness_consistency_set_expanded} indicates that the roughness constraint in this work does not independently limit the normal force or the feed speed, but instead constrains their coupled relation. Furthermore, according to Eq.~\eqref{eq:roughness_stagewise_again}, when the target roughness is fixed, the normal force and the feed speed satisfy the following implicit coupling relation:
\begin{equation}
v_t=
\left(
\frac{R_{a,t}^{\star}}{\bar{C}_{s_t}(F_t^n)^{\alpha}}
\right)^{1/\gamma}
\label{eq:implicit_coupling}
\end{equation}

However, roughness consistency alone is not sufficient, because some parameter combinations satisfying Eq.~\eqref{eq:roughness_consistency_set_expanded} may still be infeasible for the robot system. For example, the normal force may exceed the stable force-control range, the feed speed may surpass the allowable servo limit, or excessively large variations between adjacent time steps may lead to discontinuous execution. Therefore, beyond the roughness-consistency set, a physical-feasibility set is further defined as
\begin{equation}
\begin{alignedat}{2}
\Omega_{P,t}
&=
\biggl\{
(F_t^n,v_t)
\ \bigg|\ 
&&F_t^{\min}\le F_t^n\le F_t^{\max}, \\
&&&v_t^{\min}\le v_t\le v_t^{\max}, \\
&&&|F_t^n-F_{t-1}^n|\le r_F\Delta t, \\
&&&|v_t-v_{t-1}|\le r_v\Delta t
\biggr\}.
\end{alignedat}
\label{eq:physical_feasible_set}
\end{equation}
Accordingly, the final valid parameter set at a single step is written as
\begin{equation}
\Omega_t=\Omega_{R,t}\cap\Omega_{P,t}.
\label{eq:joint_feasible_set}
\end{equation}

It should be noted that the joint constraint set in Eq.~\eqref{eq:joint_feasible_set} is not a pre-defined static region. Instead, it is an online feasible set jointly determined by the target roughness $R_{a,t}^{\star}$, the stage-wise fixed spindle speed $\bar{\omega}^{(s_t)}$, the roughness model parameters calibrated offline, and the execution state at the previous time step. Therefore, the joint constraint set considered in this paper is essentially an online constraint set that varies with the task stage, trajectory location, and historical execution state.

At the $k$-th reverse denoising step of diffusion, the network first outputs a candidate action sequence $\bar{\mathbf{A}}_t^{k-1}$. For each candidate process parameter pair $(F_t^n,v_t)$ therein, the predicted roughness is first estimated online using Eq.~\eqref{eq:roughness_stagewise_again}, and the roughness-consistency energy is constructed based on the deviation from the target roughness:
\begin{equation}
\mathcal{E}_{R}(\mathbf{A}_t)
=
\sum_{\tau=t}^{t+H-1}
w_{\tau}
\left(
\log \hat{R}_{a,\tau}(F_{\tau}^n,v_{\tau})
-
\log
R_{a,\tau}^{\star}
\right)^2
\label{eq:roughness_energy}
\end{equation}
where $w_{\tau}$ denotes the importance weight of different path segments. Equation~\eqref{eq:roughness_energy} does not simply evaluate the log-space roughness error after generation is completed; instead, it continuously suppresses those samples that may appear smooth but fail to satisfy the target roughness requirement in terms of the combination of $F^n$ and $v$ during diffusion sampling.

On this basis, the physical-feasibility energy is further introduced:
\begin{equation}
\mathcal{E}_{P}(\mathbf{A}_t)
=
\mathcal{E}_{B}(\mathbf{A}_t)
+
\mathcal{E}_{D}(\mathbf{A}_t)
\label{eq:physical_energy}
\end{equation}
with the boundary constraint term
\begin{equation}
\begin{aligned}
\mathcal{E}_{B}
=
\sum_{\tau=t}^{t+H-1}
\Big(
&[F_{\tau}^n-F^{\max}]_+^2
+
[F^{\min}-F_{\tau}^n]_+^2 \\
&+
[v_{\tau}-v^{\max}]_+^2
+
[v^{\min}-v_{\tau}]_+^2
\Big).
\end{aligned}
\label{eq:boundary_energy}
\end{equation}
and the continuity constraint term
\begin{equation}
\mathcal{E}_{D}
=
\sum_{\tau=t}^{t+H-1}
\Big(
[|F_{\tau}^n-F_{\tau-1}^n|-r_F\Delta t]_+^2
+
[|v_{\tau}-v_{\tau-1}|-r_v\Delta t]_+^2
\Big)
\label{eq:continuity_energy}
\end{equation}
where $[x]_+=\max(x,0)$. Accordingly, the complete constraint energy can be written as
\begin{equation}
\mathcal{E}_{\mathrm{qual}}(\mathbf{A}_t)=\lambda_R \mathcal{E}_{R}+\lambda_P \mathcal{E}_{P}
\label{eq:total_quality_energy}
\end{equation}

Different from the common strategy of first imposing boundary limits and then checking quality afterward, $\mathcal{E}_{R}$ plays the dominant role in this work, while $\mathcal{E}_{P}$ serves as an auxiliary term. This means that, in the logic of parameter generation, the diffusion process is first required to satisfy the coupling relation prescribed by the target roughness, and only then required to satisfy robot-level executability.

At the high-noise and intermediate-noise stages, a roughness-consistency-dominant soft guidance strategy is adopted. Specifically, during reverse denoising, the gradient of the roughness-consistency energy is first used to correct the generation direction of the parameters, and a weaker feasibility correction is then superimposed:
\begin{equation}
\begin{aligned}
\mathbf{A}_t^{k-1}
&=
\bar{\mathbf{A}}_t^{k-1}
-
\gamma_k^{R}\nabla_{\mathbf{A}}
\mathcal{E}_{R}(\bar{\mathbf{A}}_t^{k-1}) \\
&\quad
-
\gamma_k^{P}\nabla_{\mathbf{A}}
\mathcal{E}_{P}(\bar{\mathbf{A}}_t^{k-1}),
\qquad k>k_c .
\end{aligned}
\label{eq:guided_denoising_update}
\end{equation}
where $\gamma_k^{R}$ and $\gamma_k^{P}$ denote the guidance coefficients for the roughness-consistency constraint and the physical-feasibility constraint, respectively. Typically, $\gamma_k^{R}>\gamma_k^{P}$ is chosen to reflect the priority of ``satisfying roughness coupling first, and physical feasibility second.''

It should be emphasized that this work does not adopt the strategy of applying a one-shot projection only after the entire diffusion process is completed. If the parameters are allowed to remain far away from the roughness-consistency set throughout denoising, the final candidate samples may deviate significantly from the feasible region. In that case, a terminal one-shot projection would inevitably require large corrections to the entire action sequence, thereby damaging the temporal structure and stage semantics learned by diffusion. In contrast, Eq.~\eqref{eq:guided_denoising_update} progressively guides the samples toward the roughness-consistency tolerance band throughout the denoising process, thus reducing the magnitude of final correction and improving the stability of the generated result.

When denoising enters the low-noise stage, the candidate parameters become close to the final output, and the final parameters must then be explicitly guaranteed to lie inside the joint constraint set $\Omega_t$. Therefore, a minimum-modification projection is performed on the candidate parameters:
\begin{equation}
(F_{\tau}^{n\star},v_{\tau}^{\star})
=
\arg\min_{F^n,v}
\left\|
\begin{bmatrix}
F^n\\
v
\end{bmatrix}
-
\begin{bmatrix}
F_{\tau}^{n(0)}\\
v_{\tau}^{(0)}
\end{bmatrix}
\right\|_2^2,
\quad
\mathrm{s.t.}\ (F^n,v)\in\Omega_{\tau}
\label{eq:projection_final}
\end{equation}
where $(F_{\tau}^{n(0)},v_{\tau}^{(0)})$ denotes the original output parameters of diffusion. The role of this projection problem is to pull the final process parameters back into the region where the roughness requirement is satisfied and physical executability is guaranteed, while modifying the original network output as little as possible. In implementation, the projection is solved in the log-transformed $(\log F^n,\log v)$ space, where the roughness-consistency band is represented as a linear strip; the boundary and rate constraints are then enforced by clipping and a small-scale quadratic projection.

Combining the stage-conditioned reverse denoising in Chapter~3, the final unified reverse update can be written as
\begin{equation}
\begin{split}
\mathbf{A}_t^{k-1}
&=
\frac{1}{\sqrt{\alpha_k}}
\left(
\mathbf{A}_t^k
-
\frac{1-\alpha_k}{\sqrt{1-\bar{\alpha}_k}}
\hat{\epsilon}_t^k
\right) \\
&\quad +
\sigma_k\xi
-
\gamma_k^{R}\nabla_{\mathbf{A}}\mathcal{E}_{R}(\mathbf{A}_t^k)
-
\gamma_k^{P}\nabla_{\mathbf{A}}\mathcal{E}_{P}(\mathbf{A}_t^k)
\end{split}
\label{eq:final_reverse_update}
\end{equation}
When $k\le k_c$, the generated feed-force pair is further projected onto the feasible set $\Omega_{\tau}$ in the late denoising steps.

Equation~\eqref{eq:final_reverse_update} together with this late-step projection indicates that the final feed-force parameters in this work are not obtained by unconstrained diffusion sampling. Instead, they are generated under the joint action of the parameter-coupling relation defined by the target roughness model and the feasible region at the robot execution level. The former guarantees roughness controllability, while the latter guarantees temporal continuity and executability of the parameters.

\begin{figure*}
    \centering
    \includegraphics[width=0.7\textwidth]{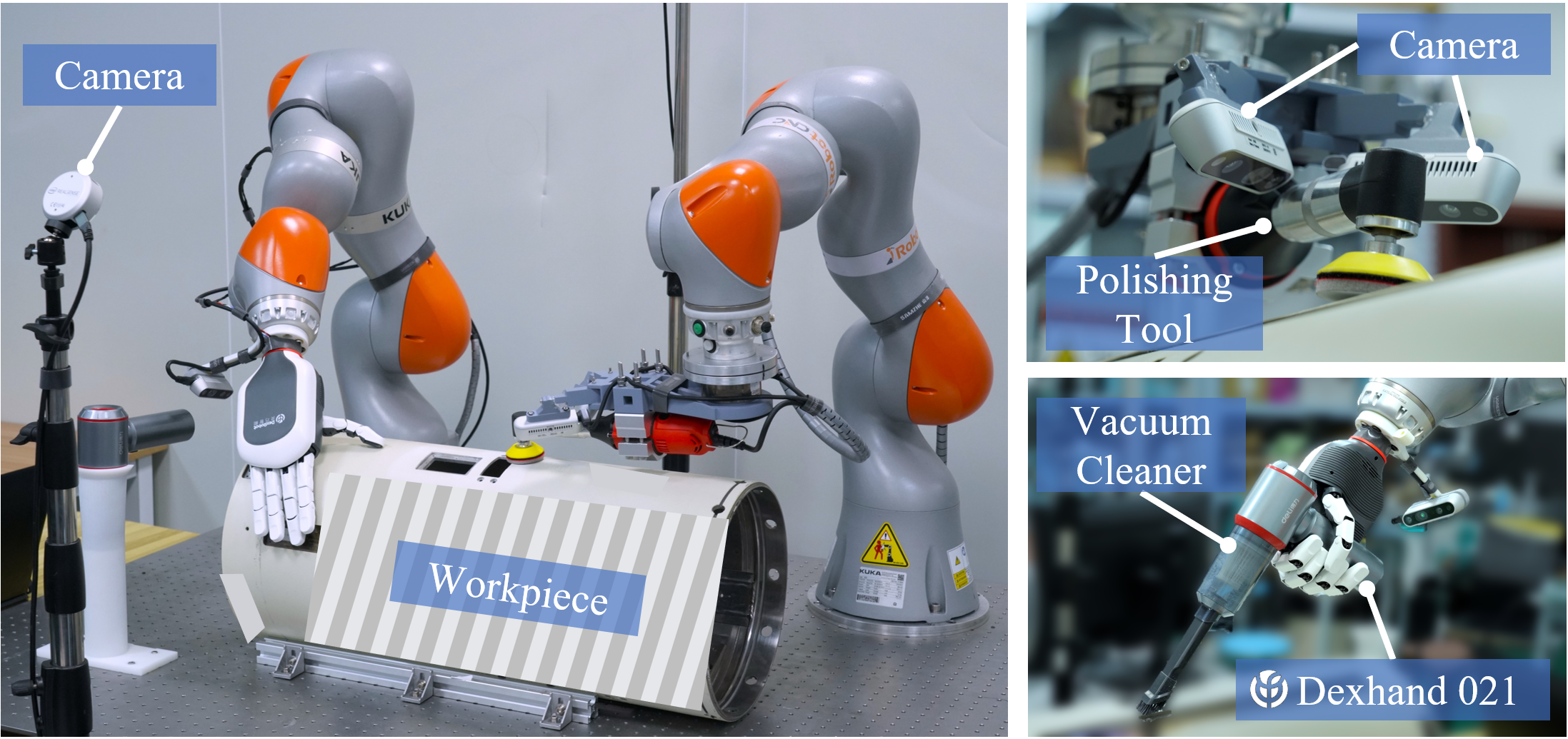}
    \caption{Dual-arm robotic polishing platform used in the experiments. The system consists of two KUKA iiwa14 robots, a polishing tool, a dexterous hand, a vacuum cleaner, and a multi-view perception system. The right arm performs polishing operations, while the left arm provides auxiliary stabilization, grasping, and handling functions. Multiple RGB-D cameras are used to capture visual observations for stage inference and action generation.}
    \label{fig:exp_setup}
\end{figure*}

\section{Experiments}
To validate the effectiveness of the proposed method in complex robotic polishing tasks, experiments are conducted on two representative scenarios, namely spacecraft cabin coating-surface polishing and inner-cavity assembly surface finishing. These two tasks simultaneously involve multi-stage process transitions, constrained contact interaction, and coupled process-parameter control, thereby imposing stringent requirements on both the long-horizon action generation capability and the process-consistency regulation ability of the policy. The proposed method is systematically evaluated in terms of task completion performance, execution stability, and surface-quality improvement.

\subsection{Experimental Setup}
To conduct the above experiments, we built a dual-arm robotic polishing platform consisting of two KUKA iiwa14 manipulators, a multi-view visual perception system, and polishing/cleaning end-effectors. During data collection, the operator completes the full polishing task through teleoperated demonstrations, while the system synchronously records visual images, robot poses, normal contact force, and tool states, together with stage labels annotated according to the task workflow. Since this work focuses on multi-stage long-horizon process tasks rather than single-trajectory reproduction, the demonstration data not only cover stable polishing segments, but also include key fragments such as workpiece holding, dual-arm transportation, and vacuum grasping. The resulting demonstration dataset therefore supports stage inference, action-sequence generation, and process-parameter-constrained modeling simultaneously.

In the real-robot experiments, the spindle speed is preset according to the corresponding process stage, while SRDP regulates the feed speed and normal contact force online under the preset spindle-speed setting. The finger motions of the dexterous hand are implemented by predefined scripts triggered by the corresponding tool-mode command, and are not directly controlled by the diffusion policy.

In this work, the observation horizon, execution horizon, and prediction horizon of the diffusion policy are set to $T_o=2$, $T_a=8$, and $T_p=16$, respectively. The visual inputs are first preprocessed by random cropping, with the image size set to $3\times230\times230$. The diffusion model is implemented using a one-dimensional convolutional architecture. During training, the Adam optimizer is adopted with a batch size of $64$, a learning rate of $1\times10^{-4}$, and a weight decay of $1\times10^{-6}$, while cosine decay is used for learning-rate scheduling. Both training and offline inference are performed with a 100-step DDPM procedure. For real-robot deployment, a 16-step DDIM sampler is used instead to reduce inference cost and improve online execution speed.

\begin{figure*}[!t]
    \centering
    \includegraphics[width=1.0\textwidth]{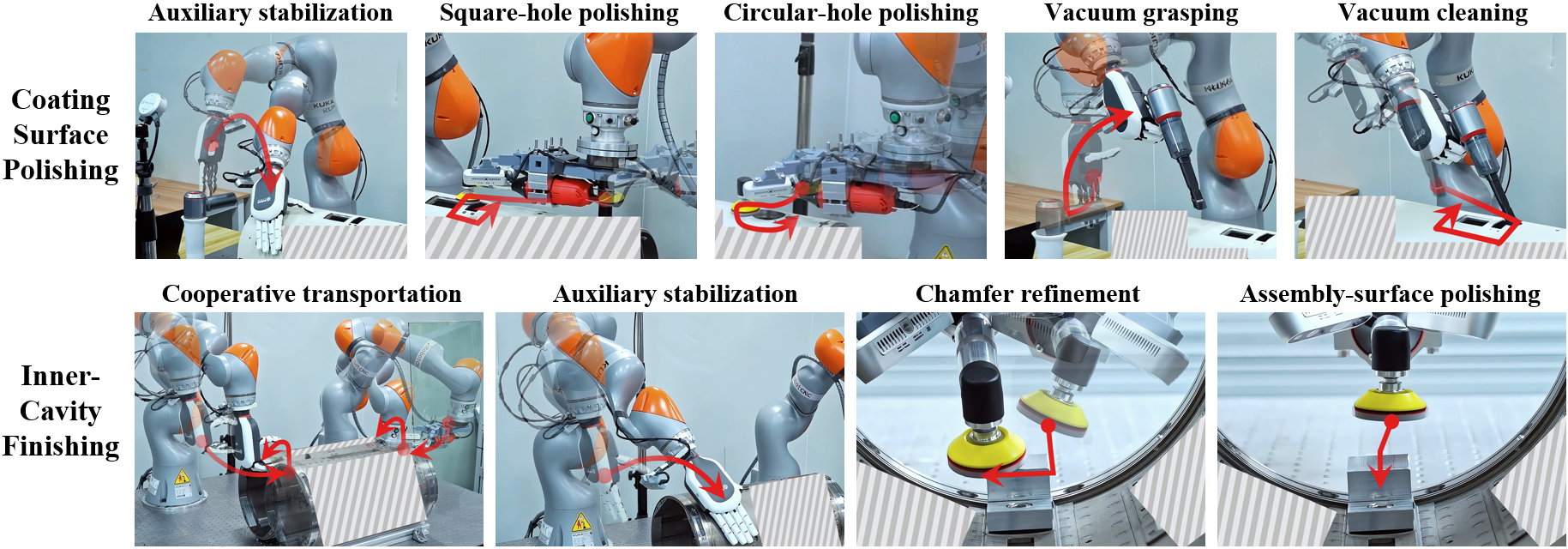}
    \caption{Representative multi-stage robotic polishing tasks used in the experiments. The coating-surface polishing task contains five stages, including auxiliary stabilization, square-hole polishing, circular-hole polishing, vacuum grasping, and vacuum cleaning. The inner-cavity finishing task contains four stages, including cooperative transportation, auxiliary stabilization, chamfer refinement, and assembly-surface polishing.}
    \label{fig:task_workflows}
\end{figure*}

\subsection{Polishing Tasks}

To evaluate the effectiveness of the proposed method in different complex polishing scenarios, two representative and mutually independent component-processing tasks are considered in this work, namely coating-surface polishing of the spacecraft cabin and inner-cavity assembly surface polishing. These two tasks differ significantly in polishing object, workspace, and process flow, and therefore provide a comprehensive testbed for evaluating the proposed method in terms of multi-stage task organization, coordinated process-parameter control, and effectiveness and generalization across scenarios.

\subsubsection{Coating-surface Polishing}

The first task is coating-surface polishing of the spacecraft cabin. This task addresses the problem of uneven coating on the sprayed outer surface of the cabin, where key regions such as frame edges need to be refined, followed by dust cleaning after polishing. According to the practical workflow, this task can be divided into five consecutive stages as follows:
\begin{itemize}[leftmargin=1.2em,itemsep=0pt,topsep=2pt,parsep=0pt,partopsep=0pt,label=\scriptsize$\bullet$]
    \item \textbf{Auxiliary stabilization:} The robot uses its left arm to stabilize the cabin so as to improve the workpiece stability during the subsequent contact-rich polishing process.
    \item \textbf{Square-hole polishing:} The robot uses the end face of the polishing tool to polish along the boundary of the square hole, thereby performing local coating refinement.
    \item \textbf{Circular-hole polishing:} The robot uses the polishing tool to polish the edge of the circular hole so as to improve the surface quality around the boundary region.
    \item \textbf{Vacuum grasping:} The robot uses the dexterous hand to grasp the vacuum cleaner in preparation for the subsequent cleaning operation.
    \item \textbf{Vacuum cleaning:} The robot moves the vacuum cleaner over the processed region in a coverage manner to remove residual dust generated during polishing.
\end{itemize}

The main difficulty of this task lies in the fact that the vacuum-cleaning stage and the polishing stages differ significantly in action semantics and execution patterns. Therefore, this task can effectively evaluate the capability of the policy in stage recognition and mode switching within a multi-stage process workflow. In addition, edge-region processing requires the robot to maintain stable contact and continuous motion within a local workspace, while the spindle speed, feed speed, and normal force must remain coordinated over a long horizon to achieve consistent processing quality.
The spindle speed is set to $2000~\mathrm{rpm}$ in this task. For this task, the roughness-model parameters calibrated from normalized process variables are set as $(C,\alpha,\beta,\gamma)=(1.00,-0.16,-0.28,0.33)$. Repeated trials were conducted on available comparable coating features, where the coating surface could be rapidly restored by thin re-spraying and local repair after polishing or failed trials.

\subsubsection{Inner-Cavity Assembly Surface Finishing}

The second task is inner-cavity assembly surface polishing of the spacecraft cabin. This task focuses on the assembly surface and chamfer regions inside the cabin, with the goal of removing residual milling marks on the assembly surface, reducing the surface roughness, and locally refining the boundary chamfer regions. Since the processing area is located inside the inner cavity, this task is more challenging than outer-surface polishing in terms of spatial reachability, contact stability, and dual-arm coordination. According to the practical workflow, this task can be divided into four consecutive stages as follows:
\begin{itemize}[leftmargin=1.2em,itemsep=0pt,topsep=2pt,parsep=0pt,partopsep=0pt,label=\scriptsize$\bullet$]
    \item \textbf{Cooperative transportation:} The two robot arms cooperatively transport the cabin to the designated workstation and establish a suitable initial pose for precise contact operations.
    \item \textbf{Auxiliary stabilization:} The robot further supports and stabilizes the cabin to improve overall stability during contact-rich processing in the confined space.
    \item \textbf{Chamfer refinement:} The robot polishes the chamfer region along the inner-cavity boundary to remove local burrs and improve the boundary transition quality.
    \item \textbf{Assembly-surface polishing:} The robot further polishes the assembly surface to reduce the surface roughness and remove residual milling marks.
\end{itemize}

The main difficulty of this task lies in the confined inner-cavity workspace, where the motion range and visual observation of the execution arm are both significantly restricted, and even slight deviations may lead to occlusion, path limitation, or contact instability. Meanwhile, although chamfer polishing and assembly-surface polishing are both material-removal processes, they differ substantially in local geometric characteristics and processing objectives. Therefore, this task not only requires stable operation in confined spaces, but also demands reliable action switching and coordinated process-parameter control across different sub-stages.
The spindle speed is set to $4000~\mathrm{rpm}$ in this task. For this task, the roughness-model parameters calibrated from normalized process variables are set as $(C,\alpha,\beta,\gamma)=(0.82,-0.11,-0.22,0.29)$. Repeated trials were conducted using a batch of replaceable assembly-boss workpieces mounted inside the cabin section, whose upper surfaces could be re-milled after each group of experiments for reuse.

\subsubsection{Failure Mode Classification}
To facilitate the subsequent analysis of experimental results, the typical failure modes in the two complex polishing tasks are further summarized, as shown in Table \ref{tab:failure_taxonomy}. According to the main modeling focus of this work, the failures are grouped into three core categories, namely stage-switching failure, contact-stability failure, and process-parameter failure, together with an additional Other category for a small number of issues that cannot be directly assigned to the first three categories.

Specifically, stage-switching failure refers to incorrect phase transition or action-semantic confusion in multi-stage process execution. Contact-stability failure refers to unstable local contact, insufficient support, or abnormal boundary following. Process-parameter failure refers to quality-related errors caused by the mismatch among normal force, feed speed, and spindle speed. In addition, a small number of issues that cannot be directly attributed based only on observable phenomena are grouped into the Other category.

\begin{figure}[]
    \centering
    \includegraphics[width=1.00\columnwidth]{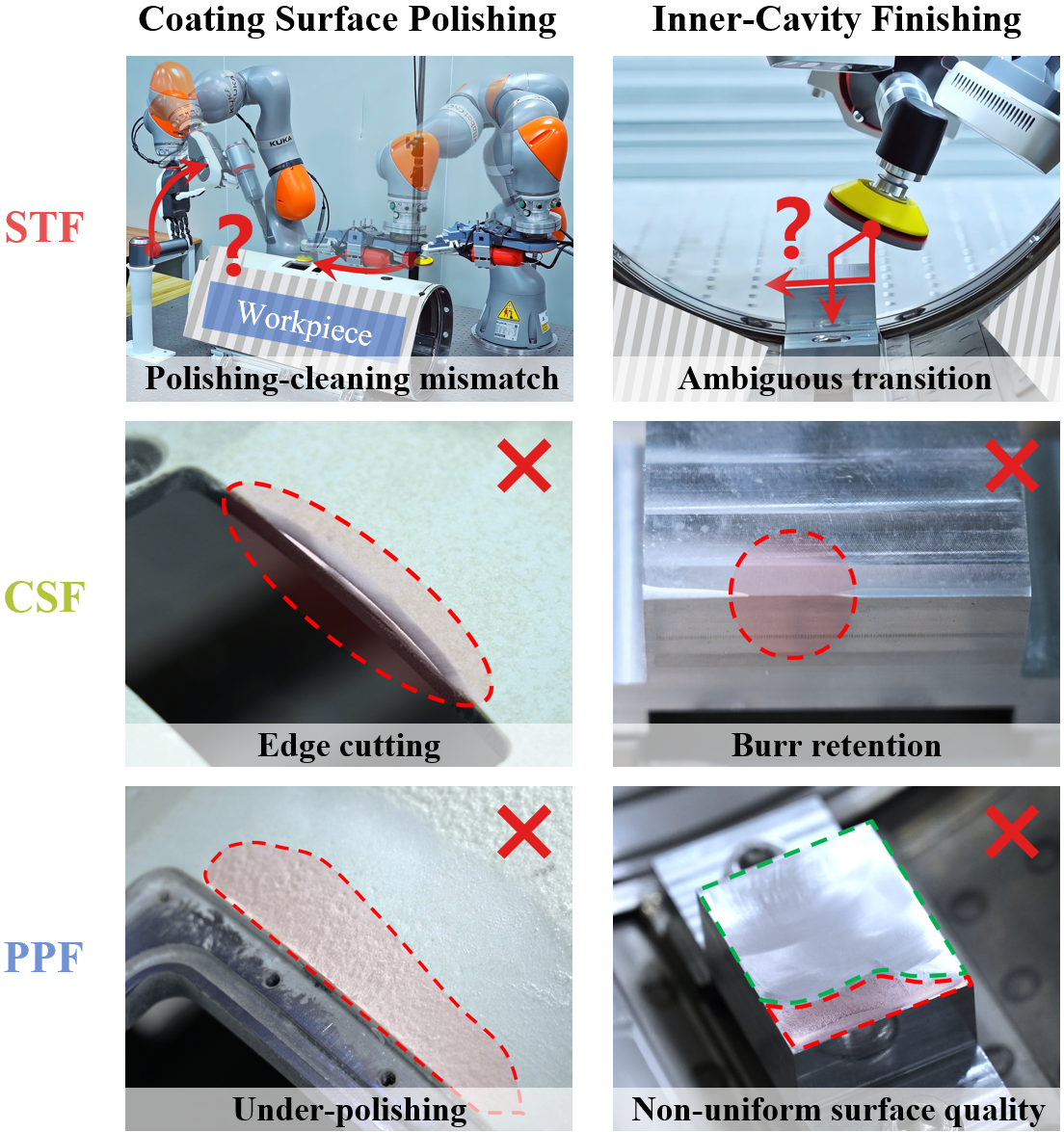}
    \caption{Representative failure cases in the two polishing tasks. The examples illustrate typical stage-switching failures, contact-stability failures, and process-parameter failures, including polishing-cleaning mismatch, ambiguous stage transition, edge cutting, burr retention, under-polishing, and non-uniform surface quality.}
    \label{fig:failure_examples}
\end{figure}

It should be noted that the classification in Table \ref{tab:failure_taxonomy} is introduced to provide a unified terminology framework for the following analysis, and some observable outcomes, although possibly influenced by multiple factors, are categorized according to the dominant mechanism. Representative visual examples corresponding to the above failure categories are further shown in Fig.~\ref{fig:failure_examples}, providing an intuitive illustration of how stage-switching, contact-stability, and process-parameter failures appear in the two polishing tasks.

\begin{table}[!t]
\centering
\fontsize{8.2pt}{9.4pt}\selectfont
\fontfamily{ptm}\selectfont
\renewcommand{\arraystretch}{1.10}
\setlength{\tabcolsep}{2.2pt}

\caption{Failure-mode classification in the two polishing tasks.}
\label{tab:failure_taxonomy}

\begin{tabular*}{\columnwidth}{@{\extracolsep{\fill}}
>{\centering\arraybackslash}m{1.05cm}
>{\raggedright\arraybackslash}m{2.95cm}
>{\raggedright\arraybackslash}m{3.15cm}@{}}
\toprule
\makecell[c]{Failure\\category}
& \makecell[c]{Coating surface\\polishing}
& \makecell[c]{Inner-cavity\\finishing} \\
\midrule

\textbf{STF}
& Mismatch between polishing and vacuum cleaning
& Ambiguous transition between chamfer refinement and surface polishing \\
\addlinespace[4pt]

\textbf{CSF}
& Unstable contact; edge cutting; support instability; incomplete dust removal
& Support instability; local contact interruption; insufficient chamfer refinement; burr retention \\
\addlinespace[4pt]

\textbf{PPF}
& Over-polishing; under-polishing; boundary-quality fluctuation
& Non-uniform surface quality; inconsistent roughness improvement \\
\addlinespace[4pt]

\textbf{OTH}
& Vacuum grasping failure
& Local collision; transport deviation \\
\bottomrule
\end{tabular*}

\vspace{1mm}
\parbox{\columnwidth}{\fontsize{7.6pt}{8.6pt}\selectfont
\textit{Abbreviations:} STF (stage-switching failure), CSF (contact-stability failure), PPF (process-parameter failure), and OTH (other hard-to-classify issues).}
\end{table}

\subsection{Comparative Experiments}
To evaluate the effectiveness of the proposed method in complex multi-stage robotic polishing tasks, SRDP is compared with three representative advanced methods, including Diffusion Policy (DP), Hierarchical Imitation Learning (HIL), and Adaptive Compliance Policy (ACP). Specifically, DP is used to represent the baseline performance without explicit stage modeling or process constraints, HIL is adopted to examine the effect of hierarchical stage organization on long-horizon task execution, and ACP is introduced to analyze the contribution of compliance control to contact stabilization. All methods are tested under the same training data, observation inputs, and execution protocol. For each method and each task, the comparative experiment is repeated in three independent groups with eight trials per group, resulting in 24 trials in total. The reported mean and standard deviation are calculated across the three groups. The Human row reports the task-completion quality of the teleoperated demonstrations used for data collection, and is provided as a reference for demonstration quality rather than as an autonomous policy baseline. For methods whose stage-transition success rate is zero, additional prompts are used during evaluation to trigger the subsequent stage. Therefore, the reported success rates of later subtasks for these methods only reflect their local execution capability, rather than their ability to autonomously complete the full long-horizon process.

\begin{table*}[!t]
\centering
\fontsize{8.4pt}{9.6pt}\selectfont
\fontfamily{ptm}\selectfont
\renewcommand{\arraystretch}{1.08}

\caption{Quantitative comparison of different methods in coating-surface polishing tasks.}
\label{tab:coating_polishing_results}
\setlength{\tabcolsep}{2.0pt}
\begin{tabular*}{\textwidth}{@{\extracolsep{\fill}}ccccccccc@{}}
\toprule
Task & Method & Metric
& \makecell[c]{Auxiliary-\\stabilization}
& \makecell[c]{Square-hole\\polishing}
& \makecell[c]{Circular-hole\\polishing}
& \makecell[c]{Vacuum\\grasping}
& \makecell[c]{Vacuum\\cleaning}
& Average \\
\midrule
\multirow{9}{*}{\makecell[c]{Coating-\\Polishing}}
& \multirow{2}{*}{DP}
& Succ. Rate($\uparrow$)
& $0.92\pm0.07$ & $0.62\pm0.00$ & $0.62\pm0.00$ & $0.75\pm0.00$ & $0.83\pm0.07$ & $0.75$ \\
&
& Transition Succ($\uparrow$)
& -- & $0.38\pm0.00$ & $0.12\pm0.00$ & $0.00\pm0.00$ & $0.00\pm0.00$ & $0.12$ \\
\addlinespace[2pt]

& \multirow{2}{*}{HIL}
& Succ. Rate($\uparrow$)
& $0.83\pm0.07$ & $0.29\pm0.07$ & $0.33\pm0.07$ & $0.62\pm0.00$ & $0.42\pm0.07$ & $0.50$ \\
&
& Transition Succ($\uparrow$)
& -- & $0.83\pm0.07$ & $0.38\pm0.00$ & $0.83\pm0.07$ & $0.58\pm0.07$ & $0.66$ \\
\addlinespace[2pt]

& \multirow{2}{*}{ACP}
& Succ. Rate($\uparrow$)
& $1.00\pm0.00$ & $0.71\pm0.07$ & $0.75\pm0.00$ & $0.75\pm0.00$ & $0.83\pm0.07$ & $0.81$ \\
&
& Transition Succ($\uparrow$)
& -- & $0.33\pm0.07$ & $0.12\pm0.00$ & $0.00\pm0.00$ & $0.00\pm0.00$ & $0.11$ \\
\addlinespace[2pt]

& \multirow{2}{*}{SRDP}
& Succ. Rate($\uparrow$)
& $1.00\pm0.00$ & $0.96\pm0.07$ & $0.88\pm0.00$ & $0.88\pm0.00$ & $1.00\pm0.00$ & $0.94$ \\
&
& Transition Succ($\uparrow$)
& -- & $0.88\pm0.00$ & $0.88\pm0.00$ & $0.88\pm0.00$ & $0.88\pm0.00$ & $0.88$ \\
\addlinespace[2pt]

& Human
& Succ. Rate($\uparrow$)
& $1.00$ & $0.96$ & $0.96$ & $1.00$ & $1.00$ & $0.98$ \\
\bottomrule
\end{tabular*}

\vspace{2mm}
\parbox{\textwidth}{\footnotesize Note: Values are reported as mean $\pm$ standard deviation across three groups; the standard deviation indicates inter-group variation and does not define the feasible range of success rates.}

\vspace{2mm}

\caption{Quantitative comparison of different methods in inner-cavity finishing tasks.}
\label{tab:inner_cavity_finishing_results}
\setlength{\tabcolsep}{3.2pt}
\begin{tabular*}{\textwidth}{@{\extracolsep{\fill}}cccccccc@{}}
\toprule
Task & Method & Metric
& \makecell[c]{Cooperative-\\transportation}
& \makecell[c]{Auxiliary-\\stabilization}
& \makecell[c]{Chamfer-\\refinement}
& \makecell[c]{Surface\\polishing}
& Average \\
\midrule
\multirow{9}{*}{\makecell[c]{Inner-Cavity\\Finishing}}
& \multirow{2}{*}{DP}
& Succ. Rate($\uparrow$)
& $0.75\pm0.00$ & $0.88\pm0.00$ & $0.58\pm0.07$ & $0.71\pm0.07$ & $0.73$ \\
&
& Transition Succ($\uparrow$)
& -- & $0.33\pm0.07$ & $0.08\pm0.07$ & $0.00\pm0.00$ & $0.14$ \\
\addlinespace[2pt]

& \multirow{2}{*}{HIL}
& Succ. Rate($\uparrow$)
& $0.50\pm0.00$ & $0.83\pm0.07$ & $0.29\pm0.07$ & $0.54\pm0.07$ & $0.54$ \\
&
& Transition Succ($\uparrow$)
& -- & $0.79\pm0.07$ & $0.83\pm0.07$ & $0.42\pm0.07$ & $0.68$ \\
\addlinespace[2pt]

& \multirow{2}{*}{ACP}
& Succ. Rate($\uparrow$)
& $0.79\pm0.07$ & $1.00\pm0.00$ & $0.79\pm0.07$ & $0.88\pm0.00$ & $0.86$ \\
&
& Transition Succ($\uparrow$)
& -- & $0.25\pm0.00$ & $0.08\pm0.07$ & $0.00\pm0.00$ & $0.11$ \\
\addlinespace[2pt]

& \multirow{2}{*}{SRDP}
& Succ. Rate($\uparrow$)
& $0.88\pm0.00$ & $1.00\pm0.00$ & $0.88\pm0.00$ & $0.88\pm0.00$ & $0.91$ \\
&
& Transition Succ($\uparrow$)
& -- & $0.88\pm0.00$ & $0.88\pm0.00$ & $0.75\pm0.00$ & $0.83$ \\
\addlinespace[2pt]

& Human
& Succ. Rate($\uparrow$)
& $0.96$ & $1.00$ & $0.96$ & $1.00$ & $0.98$ \\
\bottomrule
\end{tabular*}

\vspace{1mm}
\parbox{\textwidth}{\footnotesize Note: Values are reported as mean $\pm$ standard deviation across three groups; the standard deviation indicates inter-group variation and does not define the feasible range of success rates.}

\end{table*}

\begin{figure}[]
    \centering
    \includegraphics[width=1.00\columnwidth]{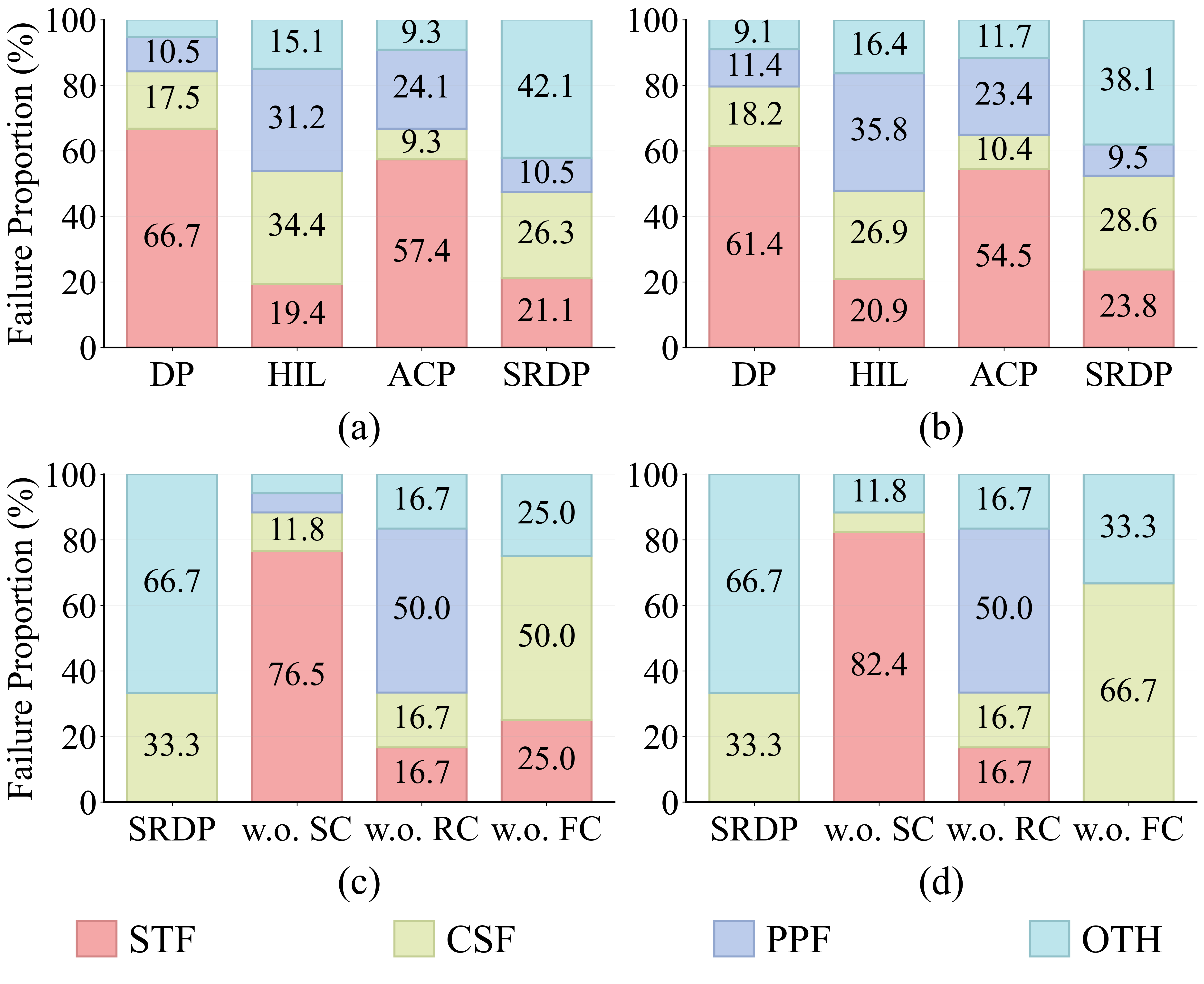}
    \caption{Failure-type distributions in comparative experiments and ablation study. Panels (a) and (b) present the comparative results on coating surface polishing and inner-cavity finishing, respectively, while panels (c) and (d) present the corresponding ablation results. For comparative experiments, each stacked bar shows the proportion of different failure categories among all failure events. For ablation study, each stacked bar shows the proportion of different dominant failure categories among failed trials. STF (stage-switching failure), CSF (contact-stability failure), PPF (process-parameter failure), and OTH (other hard-to-classify issues). DP, HIL, ACP, and SRDP denote Diffusion Policy, Hierarchical Imitation Learning, Adaptive Compliance Policy, and the proposed method, respectively. w.o.~SC, w.o.~RC, and w.o.~FC denote removing stage conditioning, roughness constraint, and feasibility/continuity constraint, respectively.}
    \label{fig:failure_type_distribution}
\end{figure}

As reported in Tables~\ref{tab:coating_polishing_results} and \ref{tab:inner_cavity_finishing_results}, SRDP achieves the best overall performance in both representative polishing tasks. For the coating-surface polishing task, SRDP obtains the highest average subtask success rate and average transition success rate (Avg.\ Succ.\ Rate = 0.94, Avg.\ Transition Succ.\ = 0.88), significantly outperforming ACP (0.81, 0.11), DP (0.75, 0.12), and HIL (0.50, 0.66). A common limitation of DP and ACP is their insufficient stage-transition capability, which is further reflected by their high proportion of stage-switching failures in Fig.~\ref{fig:failure_type_distribution}(a) (STF = 66.7\% and 57.4\%). This indicates that methods without explicit stage modeling struggle to maintain stable progression in long-horizon workflows. A further comparison shows that ACP exhibits a markedly higher proportion of process-parameter failures than DP (PPF = 24.1\% vs.\ 10.5\%), suggesting that although compliance control can partially improve local contact conditions, the lack of process constraints still leads to unreasonable parameter combinations, resulting in over-polishing, under-polishing, or boundary-quality fluctuations. HIL performs better than DP and ACP in stage transitions (Avg.\ Transition Succ.\ = 0.66), but its within-stage execution quality remains unstable, as reflected by its relatively high contact-stability and process-parameter failure ratios in Fig.~\ref{fig:failure_type_distribution}(a) (CSF = 34.4\%, PPF = 31.2\%). By contrast, SRDP not only achieves the highest overall success rate, but also significantly reduces both stage-switching failures and process-parameter failures (STF = 21.1\%, PPF = 10.5\%). Its remaining failures are mainly associated with limited contact instability and residual boundary issues (CSF = 26.3\%, OTH = 42.1\%), indicating that the proposed method effectively suppresses the dominant failure sources in the task.

A similar trend can be observed in the inner-cavity finishing task. As shown in Table~\ref{tab:inner_cavity_finishing_results}, SRDP again achieves the best results (Avg.\ Succ.\ Rate = 0.91, Avg.\ Transition Succ.\ = 0.83), outperforming ACP (0.86, 0.11), DP (0.73, 0.14), and HIL (0.54, 0.68). DP and ACP still mainly fail due to insufficient stage-transition capability, which is also reflected in Fig.~\ref{fig:failure_type_distribution}(b) (STF = 61.4\% and 54.5\%). This suggests that under the coexistence of confined spaces and multi-stage workflows, methods without explicit stage modeling continue to suffer first from unstable process progression. Although HIL substantially reduces stage-switching failures (STF = 20.9\%), its within-stage deficiencies shift further toward process-parameter mismatch and local contact instability (PPF = 35.8\%, CSF = 26.9\%). This indicates that while hierarchical scheduling improves workflow organization, it is insufficient to ensure within-stage machining quality. In contrast, SRDP maintains low stage-switching and process-parameter failure rates simultaneously (STF = 23.8\%, PPF = 9.5\%), while its remaining failures are primarily associated with contact-related and residual issues (CSF = 28.6\%, OTH = 38.1\%). Taken together, the results from both scenarios show that the advantage of SRDP lies not only in its higher overall success rate, but also in its simultaneous suppression of the two dominant failure sources, namely stage-switching failures and process-parameter failures (coating: STF/PPF = 21.1\%/10.5\%; inner: STF/PPF = 23.8\%/9.5\%).

\subsection{Ablation Study}

To analyze the contribution of each key design to the performance of complex polishing tasks, ablation studies are conducted on two representative scenarios. Considering that full multi-stage polishing experiments are time-consuming and involve relatively high material cost, the ablation study is performed on simplified workflows and evaluated under a unified protocol of three groups with six trials per group. For the coating-surface polishing scenario, only \textit{Auxiliary stabilization} and \textit{Circular-hole polishing} are retained. For the inner-cavity scenario, only \textit{Cooperative transportation}, \textit{Auxiliary stabilization}, and \textit{Assembly-surface polishing} are retained. Three ablation variants are constructed: 1) w/o Stage-Aware Modeling (w/o Stage), which removes stage inference and stage-conditioned reverse denoising; 2) w/o Roughness Constraint (w/o Rough.), which removes the roughness-consistency constraint; and 3) w/o Feasibility Constraint (w/o Feas.), which removes the parameter continuity and feasible-region constraints.

\begin{figure}[]
    \centering
    \includegraphics[width=1.00\columnwidth]{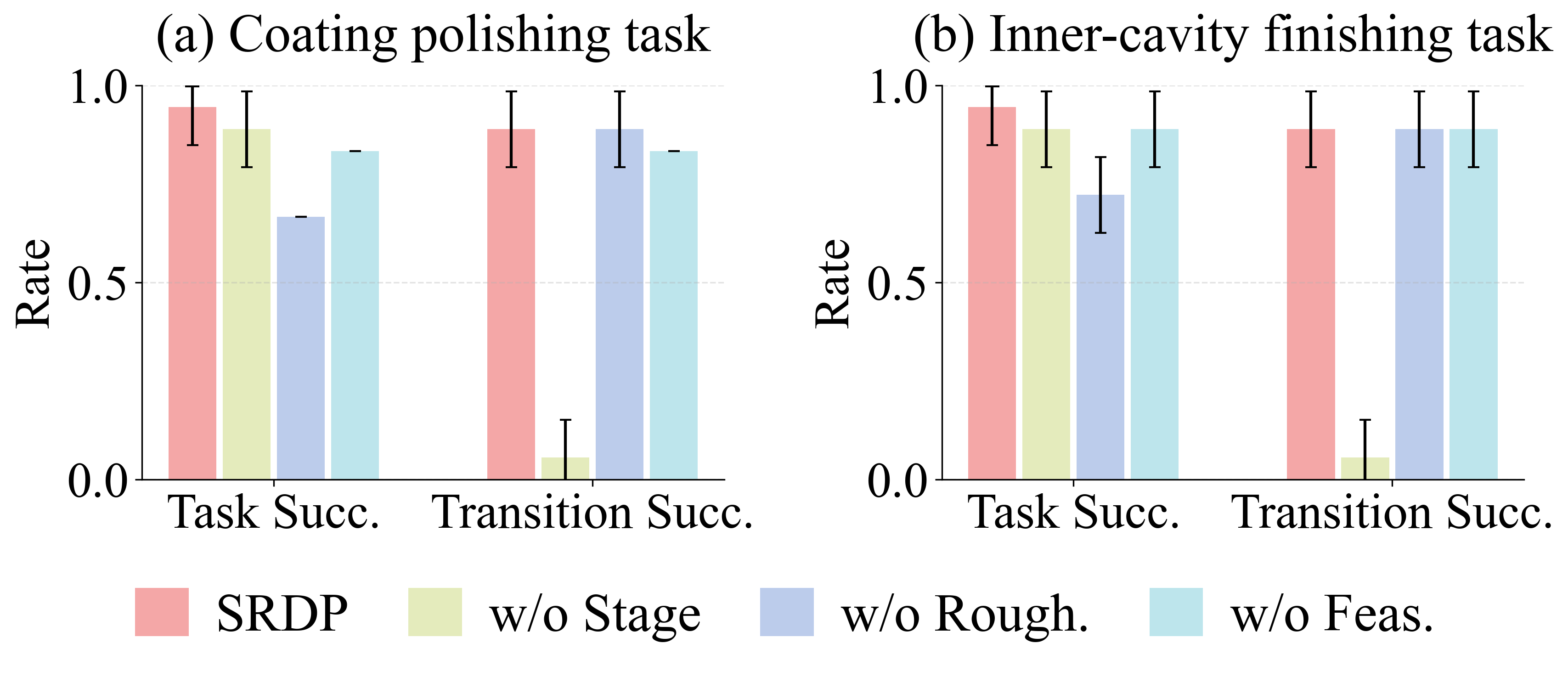}
    \caption{Ablation results on representative polishing workflows. 
    (a) Results on the simplified coating-surface polishing workflow. 
    (b) Results on the simplified inner-cavity finishing workflow. 
    Both the averaged subtask success rate and averaged stage-transition success rate are reported. Error bars indicate the standard deviation across three groups of trials.}
    \label{fig:ablation_results}
\end{figure}

As reported in Table~\ref{tab:ablation_summary} and Fig.~\ref{fig:ablation_results}, the full SRDP maintains the best overall performance in both scenarios (coating: 17/18, 16/18; inner: 17/18, 16/18). After removing stage modeling, w/o Stage shows the most severe degradation, and its transition success drops to 1/18 in both scenarios. This indicates that without stage inference and stage-conditioned denoising, the policy can no longer maintain correct process progression. Correspondingly, the dominant failures of this variant are mainly stage-switching failures, as further shown in Fig.~\ref{fig:failure_type_distribution}(c) and (d) (coating: STF = 13/17 = 76.5\%; inner: STF = 14/17 = 82.4\%). These results suggest that explicit stage-aware modeling is critical for stable long-horizon task progression.

After removing the roughness constraint, w/o Rough.\ still preserves relatively high transition success (coating/inner: 16/18), while its subtask success drops substantially (coating: 12/18; inner: 13/18). This indicates that its degradation does not mainly come from stage progression, but from within-stage processing quality. The dominant failure cause of this variant is process-parameter failure (coating: PPF = 3/6 = 50.0\%; inner: PPF = 3/6 = 50.0\%), implying that without this constraint, the policy can still reach the correct stage, but cannot reliably maintain the required coupling between normal force and feed speed under the preset spindle speed for the target surface quality.

After removing the feasibility constraint, the degradation of w/o Feas.\ lies between the above two variants (coating: 15/18, 15/18; inner: 16/18, 16/18). From the failure-type distribution, its major issue shifts to contact instability (coating: CSF = 2/4 = 50.0\%; inner: CSF = 2/3 = 66.7\%), indicating that parameter continuity and feasible-region constraints mainly contribute to stable contact maintenance and smooth physical execution. In contrast, the full SRDP only shows a few residual failures, which are mainly classified as boundary or residual issues (coating: OTH = 2/3 = 66.7\%, CSF = 1/3 = 33.3\%; inner: OTH = 2/3 = 66.7\%, CSF = 1/3 = 33.3\%). This indicates that under the complete formulation, the two dominant failure types, namely STF and PPF, have been substantially suppressed.

\begin{table}[]
\centering
\caption{Summary of ablation performance on the two representative polishing workflows.}
\label{tab:ablation_summary}
\fontsize{8.2pt}{9.2pt}\selectfont
\fontfamily{ptm}\selectfont
\renewcommand{\arraystretch}{1.05}
\setlength{\tabcolsep}{3.8pt}
\begin{tabular}{lcccc}
\toprule
\textbf{Method} 
& \makecell[c]{\textbf{Coat.}\\ \textbf{Succ. (/18)}}
& \makecell[c]{\textbf{Coat.}\\ \textbf{Trans. (/18)}}
& \makecell[c]{\textbf{Inner}\\ \textbf{Succ. (/18)}}
& \makecell[c]{\textbf{Inner}\\ \textbf{Trans. (/18)}} \\
\midrule
SRDP        & 17/18 & 16/18 & 17/18 & 16/18 \\
w/o Stage   & 16/18 & 1/18  & 16/18 & 1/18  \\
w/o Rough.  & 12/18 & 16/18 & 13/18 & 16/18 \\
w/o Feas.   & 15/18 & 15/18 & 16/18 & 16/18 \\
\bottomrule
\end{tabular}
\end{table}

Overall, the ablation results show a clear correspondence between the removed module and the dominant failure type: w/o Stage mainly increases stage-switching failures (STF = 76.5\% and 82.4\%), w/o Rough.\ mainly increases process-parameter failures (PPF = 50.0\% and 50.0\%), and w/o Feas.\ mainly increases contact-stability failures (CSF = 50.0\% and 66.7\%). This confirms that the performance gain of the full SRDP comes from the synergy of stage modeling, roughness-aware constraint, and feasibility control, rather than from any single module alone. It should also be noted that, due to experimental cost and material-consumption constraints, the ablation study is conducted on simplified workflows with a relatively small number of trials. As a result, some local variations may be unavoidable. Nevertheless, the degradation trends of different ablation variants remain consistent across the two scenarios, and still provide clear evidence of the relative role of each module.

\subsection{Machining Quality Evaluation}
Although the effectiveness of the proposed method has been validated in the previous sections from the perspectives of task success rate, stage-transition stability, and failure-type distribution, polishing is fundamentally a machining process. Therefore, it is still necessary to further evaluate the method from the perspective of final machining quality. To this end, the coating-surface roughness reduction, the assembly-surface roughness, and the chamfer width are selected as the evaluation metrics, and SRDP is further compared with ACP.

\begin{figure}[]
    \centering
    \includegraphics[width=0.98\columnwidth]{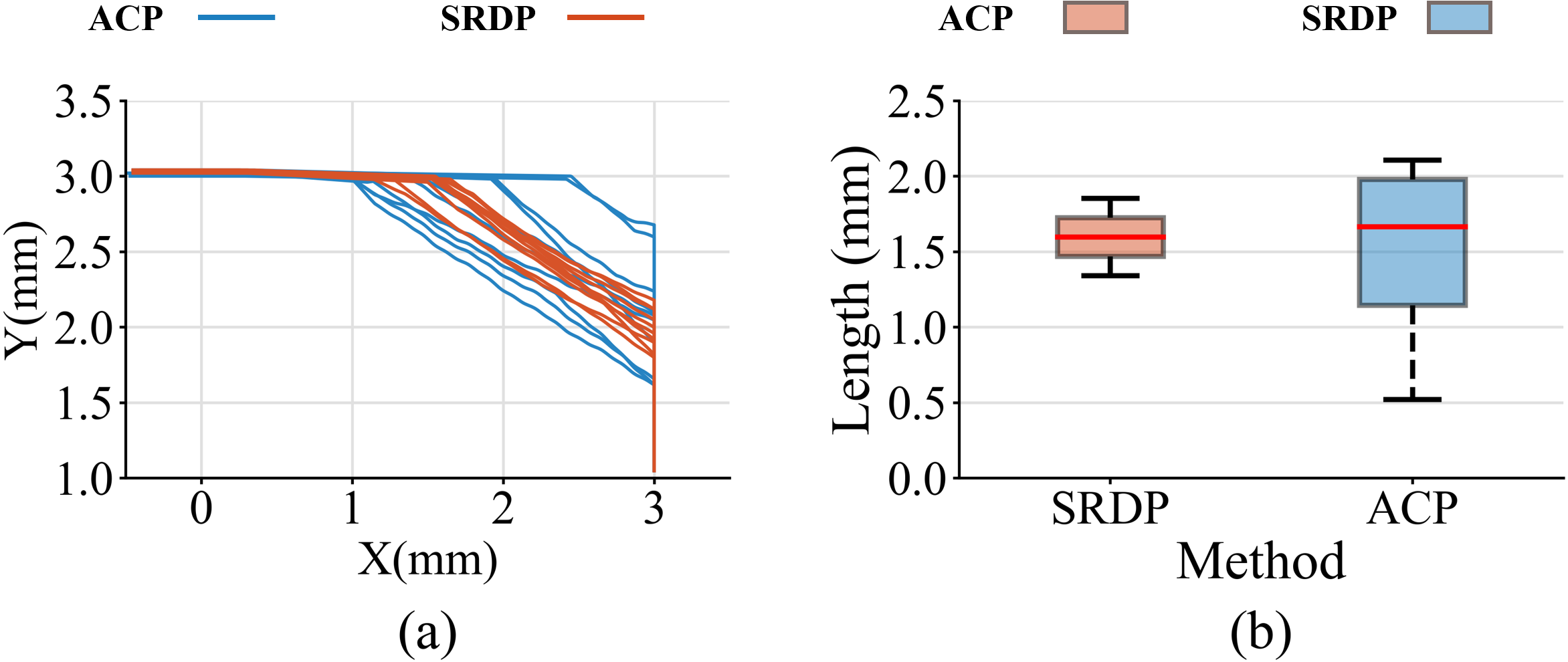}
    \caption{Chamfer-contour and chamfer-width comparison between ACP and SRDP. (a) Chamfer contours measured at ten sampling locations for the two methods. (b) Statistical distribution of chamfer widths obtained by ACP and SRDP.}
    \label{fig:chamfer_comparison}
\end{figure}

\begin{figure}[]
    \centering
    \includegraphics[width=0.98\columnwidth]{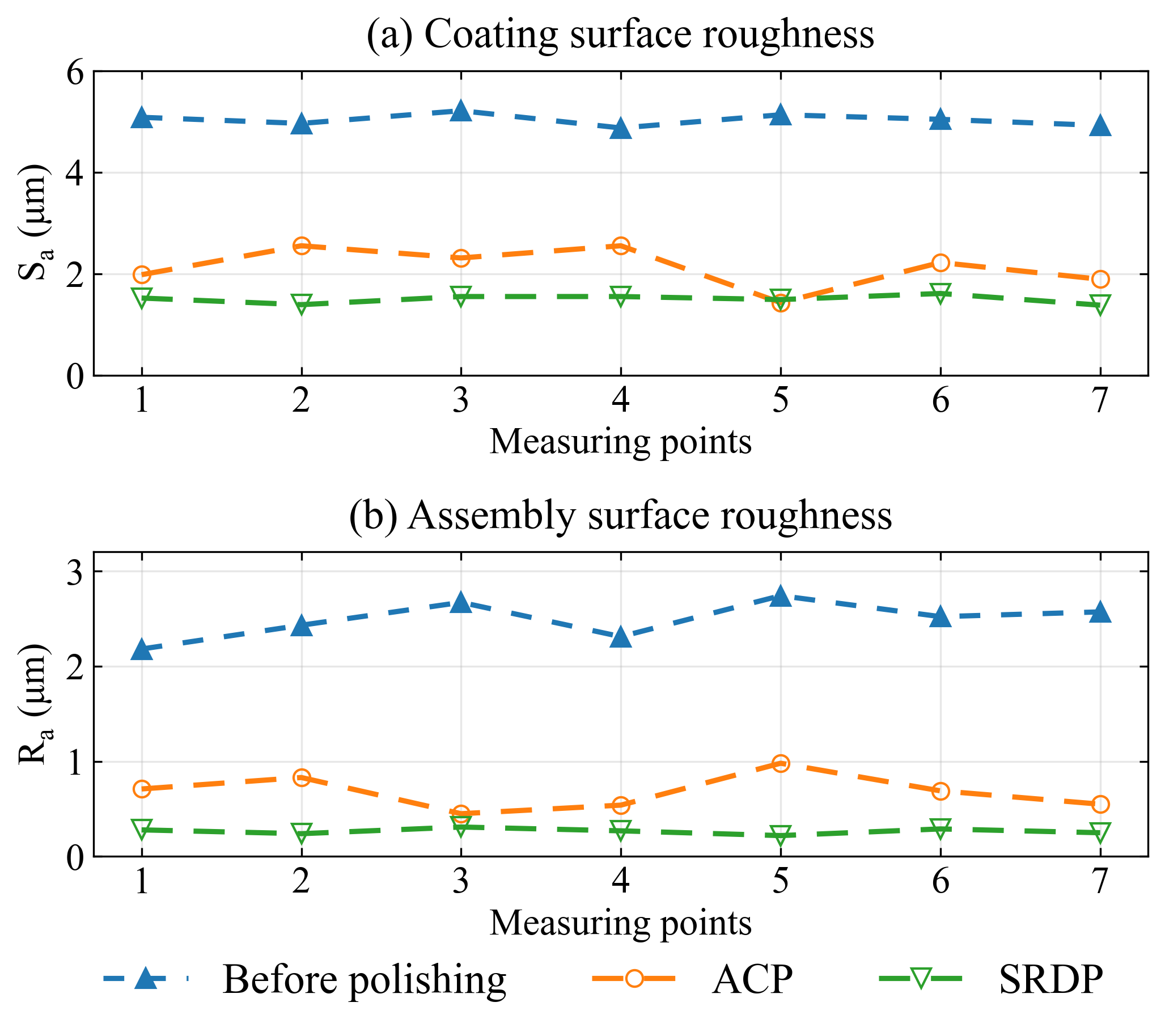}
    \caption{Roughness comparison results between ACP and SRDP. (a) Variation of coating-surface roughness, where the roughness values were obtained by white-light interferometry. (b) Variation of assembly-surface roughness, where the roughness values were measured by a surface roughness tester.}
    \label{fig:roughness_comparison}
\end{figure}

As shown in Figs.~\ref{fig:roughness_comparison} and \ref{fig:chamfer_comparison}, SRDP achieves lower and more stable roughness values at multiple measurement points in both scenarios, while also exhibiting better consistency in chamfer contours and width distribution. For the coating surface, although ACP can significantly reduce the roughness, noticeable fluctuation still remains across different measurement points. In contrast, SRDP maintains lower roughness values and better point-wise consistency throughout. For the inner-cavity assembly surface, SRDP likewise achieves superior results, with the roughness further stabilized in a lower range. Meanwhile, the chamfer contours produced by SRDP are more concentrated, and the statistical distribution of chamfer width is also more compact, indicating more consistent local geometric finishing quality. Overall, the proposed method not only improves surface quality effectively, but also enhances machining consistency, demonstrating its promising engineering applicability in real polishing tasks.

\section{Conclusion}
This paper presented SRDP, a Stage-Aware and Roughness-Constrained Diffusion Policy for complex multi-stage robotic polishing. The framework targets two key limitations of applying diffusion-based imitation learning to industrial polishing, namely long-horizon process-stage uncertainty and unconstrained process-parameter drift. SRDP infers the process-stage posterior from multimodal observation histories and uses it to condition the shared reverse denoising process, enabling stage-consistent action generation without externally provided stage labels during execution. It further incorporates roughness-oriented process constraints into diffusion sampling, where feed speed and normal contact force are constrained under stage-wise preset spindle speeds to improve process consistency and physical feasibility.

Experiments on coating-surface polishing and inner-cavity finishing validated the effectiveness of SRDP. Comparative experiments, ablation studies, and machining-quality evaluations show that SRDP improves task success and stage-transition stability while reducing the two dominant failure sources, namely stage-switching failures and process-parameter failures. The machining results further indicate that these improvements lead to lower roughness, better chamfer consistency, and more stable final surface quality. Future work will extend SRDP toward more flexible task transitions and richer manufacturing-quality objectives.

\section*{CRediT authorship contribution statement}
Shuai Ke: Writing--original draft, Methodology, Validation. Jiexin Zhang: Methodology, Writing--review \& editing. Huan Zhao: Methodology, Writing--review \& editing. Zhiao Wei: Methodology, Validation. Yikun Guo: Software, Validation. Tiange Wu: Validation, Visualization. Guoqiang Guo: Resources, Project administration. Haoyuan Zhou: Software, Validation. Jie Pan: Validation, Visualization, Writing--original draft. Han Ding: Writing--review \& editing, Resources, Project administration.

\section*{Acknowledgement}
This work was supported by the Fundamental and Interdisciplinary Disciplines Breakthrough Plan of the Ministry of Education of China under Grant No. JYB2025XDXM208, and by the National Natural Science Foundation of China under Grant Nos. U24A20130 and 52505016.

\printcredits

\bibliographystyle{unsrt} 

\bibliography{bib.bib}

\bio{}
\endbio

\endbio

\end{document}